%% file: main.tex
\documentclass[10pt,twocolumn,letterpaper]{article}

\usepackage{cvpr}
\usepackage{times}

\definecolor{cvprblue}{rgb}{0.21,0.49,0.74}
\usepackage[pagebackref,breaklinks,colorlinks,allcolors=cvprblue]{hyperref}

\input{math_commands.tex}

\usepackage{hyperref}
\usepackage{url}
\usepackage{graphicx}
\usepackage{multirow}
\usepackage{mathtools}
\usepackage{sidecap}
\usepackage{mdframed}

\input{preamble}

\title{What's in the Image? A Deep-Dive into the Vision of Vision Language Models} 

\addtocontents{toc}{\protect\setcounter{tocdepth}{0}}

\author{
Omri Kaduri\raisebox{0.5ex}{\small *}\qquad Shai Bagon\raisebox{0.5ex}{\small *}\qquad Tali Dekel\\
Weizmann Institute of Science\\
\small \raisebox{0.5ex}{\small *}Indicates equal contribution. \\
\small Project webpage: \href{https://vision-of-vlm.github.io/}{\texttt{vision-of-vlm.github.io}}}

\begin{document}

\maketitle
\begin{abstract}
Vision-Language Models (VLMs) have recently demonstrated remarkable capabilities in comprehending complex visual content. 
However, the mechanisms underlying how VLMs process visual information remain largely unexplored. 
In this paper,  we conduct a thorough empirical analysis, focusing on the attention modules across layers.
We reveal several key insights about how these models process visual data: (i)~the internal representation of the query tokens (e.g., representations of "describe the image"), is utilized by 
VLMs to store global image information; we demonstrate that these models generate surprisingly descriptive responses solely from these tokens, without direct access to image tokens.  (ii)~Cross-modal information flow is predominantly influenced by the middle layers (approximately 25\% of all layers), while early and late layers contribute only marginally. 
(iii)~Fine-grained visual attributes and object details are directly extracted from image tokens in a spatially localized manner, i.e., the generated tokens associated with a specific object or attribute attend strongly to their corresponding regions in the image.  
We propose novel quantitative evaluation to validate our observations, leveraging real-world complex visual scenes. Finally, we demonstrate the potential of our findings in facilitating efficient visual processing in state-of-the-art VLMs.
\end{abstract}

\vspace{-20pt} 
\section{Introduction}

Vision-Language Models (VLMs) have recently emerged as a powerful extension of Large Language Models (LLMs). As demonstrated by their unprecedented capabilities in generating highly detailed and accurate descriptions of complex visual scenes, these models are quickly narrowing the gap between machine-generated and human interpretation of the visual world~\cite{llava1,llava1_5,chen2024internvl,internvl1_5,achiam2023gpt4}.  As such, 
VLMs have been rapidly adopted across diverse visual tasks, including in robotics, medical imaging analysis, autonomous driving, and content generation \cite{liu2024ok, liu2023clip, tian2024drivevlm, lian2023llm}.

Despite their growing adoption, VLMs are often treated as black-box tools or agents for solving specific tasks~\cite{shaham2024multimodal, gu2024conceptgraphs}, with limited understanding of their internal mechanisms for processing visual data.  Uncovering these mechanisms is essential for enhancing model transparency, efficiency, and trustworthiness in high-stakes applications, as well as for guiding future VLM design. In this work, we take significant steps toward unraveling the ``vision'' of prominent VLMs, offering new insights into how these models interpret and process visual data.

We examine the scenario in which the VLM receives an input image along with the query ``describe the image''. The VLM generates its response autoregressively,  where each generated token gathers information from both the input image and text.  In this work, we aim to understand the information flow between the visual and textual modalities. Our analysis focuses on the attention modules across the VLM's layers through a set of experiments in which we restrict in different ways the access to visual information across layers. This allows us to uncover several critical insights: (i)~The models compress high-level \emph{image} information into the query \emph{text} tokens.
We demonstrate this insight by blocking the direct influence of image tokens on the generated tokens,
allowing visual information to be accessible only indirectly through the query text tokens.
Remarkably, the model generates descriptive responses, relying solely on the visual information encoded in the query text tokens. (ii)~Middle layers  play a crucial role in the vision-to-language knowledge transfer, 
while early and late layers contribute only marginally; we show that accessing image tokens only in mid-layers ($\sim\!25\%$ of all layers) results in minor degradation in the VLM's performance.
(iii)~Fine-grained object details and visual attributes are directly retrieved  from image tokens in a spatially localized manner. 

A key aspect in our study involves validating our observations by measuring the alignment between the VLM's original output and its modified output under each of our experiments above. This evaluation requires comparing two free-text paragraphs -- a challenging task due to possible large variations in wording and writing style. Inspired by~\cite{zheng2023judging}, we propose a new LLM-based evaluation protocol which enables us to quantify the agreement between the modified response and the original one. We ground our proposed evaluation with a human study, validating its robustness and accuracy. We further propose a novel automatic evaluation that harnesses off-the-shelf object segmentation tools \cite{kirillov2023sam} to quantitatively evaluate the emerged spatial localization across the VLM's layers.

Finally, we demonstrate that our observations facilitate efficient 
processing,  allowing to distill the VLM's internal representation into a \emph{compressed context} space. This gives rise to a new application we term ``Image Re-prompting'', which allows to efficiently ask several questions on an image, using only the compressed context. While the compressed context is $\times20$ smaller than the full 
one,
it achieves 96\% of the performance in visual question answering~\cite{fu2023mme}. 

\noindent In summary, the contributions of our work are as follows:
\begin{itemize}
    \item We reveal the surprising role of query tokens as high-level image descriptors, the critical role of the middle layers, and the 
    way by which fine-grained details are retrieved.
    \item We propose new automatic evaluation protocols that harness the use of LLMs and image segmentation tools.
    \item  We take a first step towards leveraging our understanding for efficient visual processing in VLMs.
    \item To the best of our knowledge, we are the first to consider VLMs at the scale of 76B-parameter~\cite{internvl1_5} in the context of interpretability. We further analyze \llava-7B~\cite{llava1_5}.

\end{itemize}

\section{Related Work}

\myparagraph{Vision-Language Models (VLMs).} VLMs extend Large Language Models (LLMs) to jointly process visual and textual inputs, with the LLM handling most of the computational analysis. VLMs generally consist of a pre-trained LLM, a vision encoder, and an adapter that aligns visual representations with the LLM’s embedding space. Prominent open-source VLMs build on high-performance LLMs like Llama3~\cite{dubey2024llama}, Mistral~\cite{jiang2023mistral}, and Qwen~\cite{yang2024qwen2}. Earlier VLMs leveraged CLIP~\cite{clip} as the vision encoder, while newer models employ larger encoders~\cite{internvl1_5} to handle images at various resolutions~\cite{llava1_5, internvl1_5, tong2024cambrian}. Visual embeddings are adapted to LLM space through an adapter~\cite{llava1, alayrac2022flamingo, li2023blip, tong2024cambrian}, with the MLP-based \cite{llava1} adapter is commonly used. In our study,  we analyze the state-of-the-art open-source VLM \internvl-76B~\cite{internvl1_5} and validate our findings on the widely-used \llava-7B~\cite{llava1_5} to generalize across architectures.
 
\myparagraph{Interpreting LLMs.}  
As LLMs have become widely used, interpretability research has emerged to understand various model components, including attention layers~\cite{clark2019does, chefer2021generic}, feed-forward layers~\cite{geva2020transformer}, and activation patterns~\cite{meng2022locating, conmy2023towards}. Techniques like the \emph{logit lens}~\cite{nostalgebraist2020logitlens} reveal \emph{what} information is encoded at each layer. In contrast, our work examines \emph{how} information flows through the model, analyzing attention patterns with the \emph{attention knockout} tool~\cite{geva2023dissecting}, which 
blocks 
specific 
attention connections to isolate their roles.

\myparagraph{Interpreting VLMs.}
 VLMs interpretability field is evolving, several works focused on where information is stored inside the model \cite{basu2024understanding}, revealing shortcomings of using pre-trained vision encoder \cite{tong2024eyes}, and exploring VLM hallucinations \cite{bai2024hallucination}.
Concurrent works focused on spatial localization in VLMs by employing the logit lens ~\cite{jiang2024interpreting, neo2024towards}, demonstrating that image tokens can be directly mapped to semantically-relevant words in the vocabulary.  
Our work provides a broader examination of visual processing in VLMs showing that the internal visual representation is composed of a compressed representation which provides high level information, and localized retrieval of fine-grained information. These two pathways motivate us, for the first time, to explore applications of efficient VLMs.

\myparagraph{LLM-as-a-judge}
As LLMs continue to evolve, they are increasingly recognized as viable alternatives to human annotators, allowing for scalable and reproducible assessment methods across various  tasks \cite{gilardi2023chatgpt,huang2023chatgpt}. 
In \cite{zheng2023judging}, the concept of  \emph{LLM-as-a-judge} has been established by demonstrating the use of LLMs to evaluate responses generated by other LLMs. Inspired by this concept, we introduce an LLM-based evaluation protocol for comparing two free-text image descriptions. Our approach automatically assesses object identification and detects hallucinations, offering a novel solution for evaluating image captioning accuracy. 

\section{Preliminary}
\label{sec:preliminary}
The prevalent design of VLMs includes three main components: 
a pre-trained decoder-only LLM, a pre-trained vision encoder, and an adapter~\cite{llava1,internvl1_5,chen2024internvl}. The vision encoder processes an input image by dividing it into patches, each of which is embedded into a vector. The adapter then projects these embeddings into the LLM’s token embedding space. This design allows VLMs to handle tokens from both modalities in a single sequence.
We distinguish between three types of tokens within the VLM framework, where each token type corresponds to a specific set of indices within $\textbf{T}$, the full sequence:

\begin{enumerate}
    \item \textbf{Image tokens}, \( \mathbf{T}_{\text{img}} \): Token embeddings encoding the input image. Formally, the vision encoder and adapter jointly transform the image $I$ into a set of image tokens, \( \mathbf{T}_{\text{img}} \in \mathbb{R}^{|P_{\text{img}}| \times d} \), where \( P_{\text{img}} \) denotes the indices of image tokens within the full token sequence, and \( d \) is the embedding dimension.

    \item \textbf{Query tokens}, \( \mathbf{T}_{\text{txt}} \): token embeddings of the input query text (e.g., ``describe this image”). These tokens are represented by \( \mathbf{T}_{\text{txt}} \in \mathbb{R}^{|P_{\text{txt}}| \times d} \), where \( P_{\text{txt}} \) are the indices corresponding to query tokens within the sequence.

    \item \textbf{Generated tokens}, \( \mathbf{T}_{\text{gen}} \): token embeddings of the VLM's generated response. Generated tokens expand the sequence, and at the \( i \)-th generation step, the cumulative set of generated tokens is \( \mathbf{T}_{\text{gen}}^{(i)} \in \mathbb{R}^{|P_{\text{gen}}^{(i)}| \times d} \), where \( P_{\text{gen}}^{(i)} \) represents the indices of generated tokens in the sequence up to the \( i \)-th step.
\end{enumerate}
The full token sequence processed by the VLM is then: 
\[
\mathbf{T} = [\mathbf{T}_{\text{img}}, \mathbf{T}_{\text{txt}}, \mathbf{T}_{\text{gen}}]
\]
To generate the \( (i+1) \)-th token in this sequence, the VLM
processes all tokens through a series of transformer blocks, each comprising normalization, causal self-attention, and feed-forward MLP modules. The generation relies on a causal attention mechanism, controlled by an attention mask that ensures each token attends only to previous tokens in the sequence. 
Formally, the attention is masked by $\mathbf{M}^{(i)} \in {N_i \times N_i}$, where \( N_i = |P_{\text{img}}| + |P_{\text{txt}}| + |P_{\text{gen}}^{(i)}| \), designed to enforce causality:
{\small \begin{equation}
\mathbf{M}^{(i)}[p, q] = 
\begin{cases} 
0 & \text{if } q \leq p \\
-\infty & \text{otherwise}
\end{cases}
\label{eq:causal_mask}
\end{equation}
Thus, the attention scores are computed as:
\begin{equation}
\mathbf{A} = \texttt{Att}(\mathbf{Q}, \mathbf{K}, \mathbf{M}) = \texttt{softmax}\left( \frac{\mathbf{Q} \mathbf{K}^\top}{\sqrt{d}} + \mathbf{M} \right)
\label{eq:masked_att}
\end{equation}}

\begin{figure*}[ht]
    \centering
     \includegraphics[width=1\linewidth]{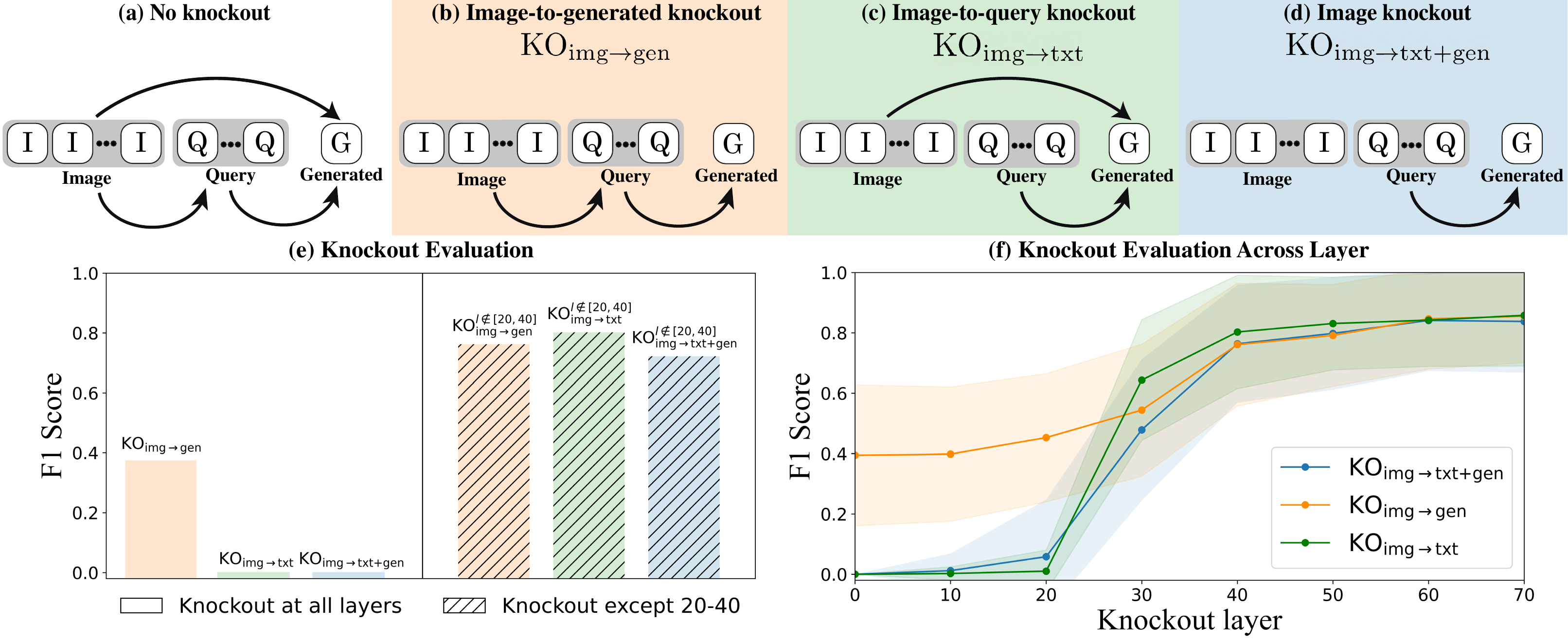} 
     \vspace{-0.8cm}     
     \caption{ \textbf{Analyzing  visual information flow via attention knockout:} (a) The VLM employs causal masking (Eq.~\ref{eq:causal_mask}), allowing generated and query tokens to gather information from image tokens, but not vice versa.  We analyze three knockout configurations: (b)~Image-to-generated $\KO{img}{gen}$: visual information flows to generated tokens only through query tokens, (c)~Image-to-query $\KO{img}{txt}$: blocks query tokens from accessing image information, and (d)~Image-to-others $\KO{img}{txt+gen}$: blocks image tokens from affecting all other tokens.
     (e) Evaluation of model responses (see Sec.~\ref{sec:llmjudge}) under each knockout configuration reveals that $\KO{img}{gen}$ achieves a 0.4 F1 score despite indirect image access, while $\KO{img}{txt}$ fails completely, demonstrating query tokens' essential role as global image descriptors. 
     (f)   We expand previous experiments by knocking out attention, starting from layer $l$. Results highlight a consistent drastic rise  in F1 scores in the mid-layers, suggesting their critical role in visual information processing. See \llava  results in Fig.~\ref{fig:knockout-sm}.}
     \label{fig:knockout}
     \afterfigure
\end{figure*}

In the sequence $\mathbf{T}$, image tokens precede query and generated tokens, and the causal mask $\mathbf{M}$ ensures that information flows from: image to query, image to generated, and query to generated tokens, as illustrated in Fig.~\ref{fig:knockout}(a). At the initial autoregressive decoding step (i=0), only image and query tokens are processed, as no tokens have yet been generated. After this step, due to causal masking, the internal representations of image and query tokens remain fixed, serving as a constant context for subsequent decoding steps.

\section{What's In The Image?}
\label{sec:analysis}

\begin{figure}
     \centering
     \includegraphics[width=1\linewidth]{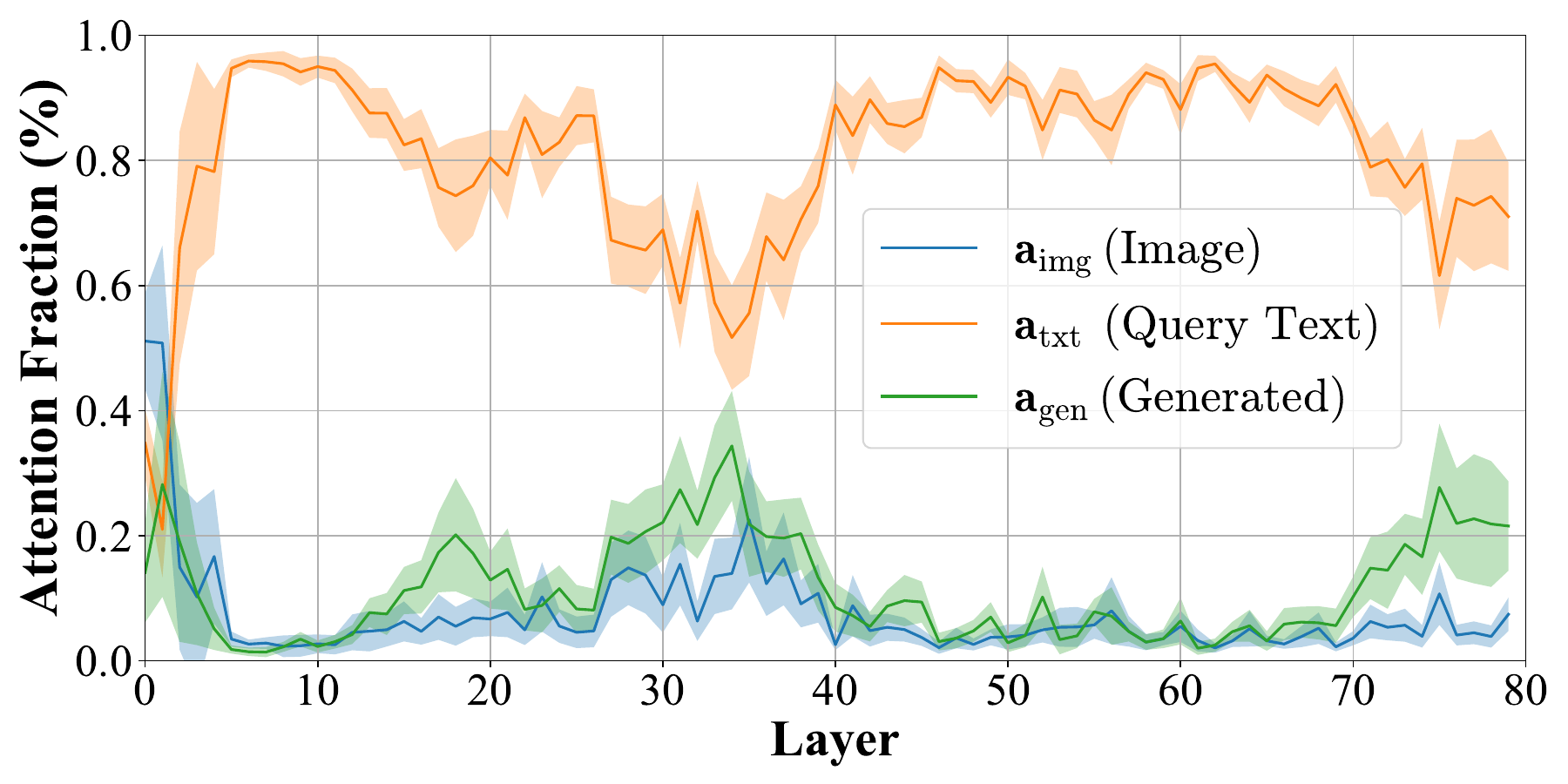} \vspace{-0.7cm}
     \caption{ \textbf{Fraction of attention to different token types:} We measure the relative amount by which the generated tokens attend to: image tokens (blue), query text tokens (orange), and the previously generated tokens in the sequence (green). We report the distribution of relative attention for a set of 80 images, averaged across attention heads and generated tokens, for \internvl-76B \cite{internvl1_5}; see Fig.~\ref{fig:llava_attention_across_layers} for results on \llava.
     }
\label{fig:attention_across_layers}
 \end{figure}
 
We consider the general task of image description, where the VLM is given an input image and is prompted with a basic instruction \emph{``describe the image''}. Our goal is to gain a better understanding of the internal mechanism by which the model leverages visual information during the autoregressive generation of its response. We provide results for \internvl-76B~\cite{internvl1_5}, a state-of-the-art large VLM, and report additional results on \llava-7B~\cite{llava1_5} in Sec.~\ref{sec:llava_results}.

Our analysis focuses on the attention modules, which govern the flow of information between the visual and textual modalities. We begin our exploration by extracting the attention values of each generated token relative to all other tokens in $\mathbf{T}$, which consists of image tokens ($\Tv$), query tokens ($\Tt$), and the previously generated tokens ($\Tg^{(i)}$).
    
The first question we raise is \emph{to what extent is the generated token influenced by the different types of tokens across layers?} 
     We quantify, per layer, the influence of each token type on the $i^{th}$ generated token by computing the relative attention directed to each type: $\mathbf{a}^{(i)}_\text{img}$, $\mathbf{a}^{(i)}_\text{txt}$, and $\mathbf{a}^{(i)}_\text{gen}$. Note that $\mathbf{a}^{(i)}_\text{img}+\mathbf{a}^{(i)}_\text{txt}+\mathbf{a}^{(i)}_\text{gen}=1$. We denote by $\mathbf{a}_\text{img}$, $\mathbf{a}_\text{txt}$, and $\mathbf{a}_\text{gen}$ the average of relative attention across generated tokens.

Figure \ref{fig:attention_across_layers}  shows the distribution of $\mathbf{a}_\text{img}, \mathbf{a}_\text{txt}, \mathbf{a}_\text{gen}$ for a set of random 80 images from COCO across layers of the VLM.
The plot reveals a non-uniform flow of information  across layers: 
$\mathbf{a}_\text{img}$ is prominent in the first few layers (0-5), then drastically drops while exhibiting a moderate increase in mid-layers (20-40). Furthermore, the majority of the attention of the generated token is directed to the embeddings of query text tokens after the very first few layers ($\mathbf{a}_\text{txt}$). 
This behavior is surprising, as the information essential for describing the image resides in the image tokens, while the input query text is generic. Moreover, despite the query tokens constituting less than 5\% of the total tokens, they command over 60\% of the overall attention.

Intrigued by these non-uniform patterns, we conduct a thorough empirical analysis to better understand the information accumulated in image and query tokens, and their roles in the generation process. Specifically, in our analysis, we knock out the information flow between different token types and evaluate the impact on the generated output. 
To this end, we propose a new LLM-based evaluation protocol, which allows us to automatically quantify the level of fidelity of the response from the VLM under knockout relative to the original response without knockout. 
Next, we describe in detail our empirical analysis and evaluation.

\subsection{Attention Knockout in VLMs}
\label{sec:knockout}
Our analysis revolves around blocking the information flow between the image tokens, to other tokens, as illustrated in Figure~\ref{fig:knockout}(a-d).
In practice, this is achieved by knocking out the attention from image tokens to either the query token ($\KO{img}{txt}$), generated tokens  ($\KO{img}{gen}$), or both ($\KO{img}{txt+gen}$). It allows us to reveal how visual information gets processed, as we will demonstrate in this section.

Formally, the general definition of the attention knockout mask, $\mathbf{M}_\text{ko}$, is given by:
{\small \begin{equation}
\mathbf{M}_\text{ko}[p, q; P_\text{src}, P_\text{tgt}] = 
\begin{cases} 
-\infty & \text{if } q \in P_\text{src} \text{\;and\;} p \in P_\text{tgt} \\
0 & \text{otherwise}
\end{cases}
\label{eq:attn_ko}
\end{equation}}
where $P_\text{src}$ and $P_\text{tgt}$ are the sets of token indices from which attention is blocked. The knockout mask is applied starting from layer $l$, with earlier layers remaining unaffected and only subject to the causal mask. We varied the value of $l$ to examine the impact of blocking attention from different layers.

For each experiment, specific sets $P_\text{src}$ and $P_\text{tgt}$ were used to define distinct knockout configurations, detailed in the experimental setup.
We add the knockout mask to the causal mask (Eq.~\ref{eq:causal_mask}), updating the attention scores computation as:
{\small \begin{equation}
\texttt{Att}(\mathbf{Q}, \mathbf{K}, \mathbf{M}) = \texttt{softmax}\left( \frac{\mathbf{Q} \mathbf{K}^\top}{\sqrt{d}} + \mathbf{M} + \mathbf{M^\text{ko}}\right)
\label{eq:masked_ko_att}
\end{equation}}

\begin{SCfigure}[80]
    \includegraphics[width=.68\linewidth]{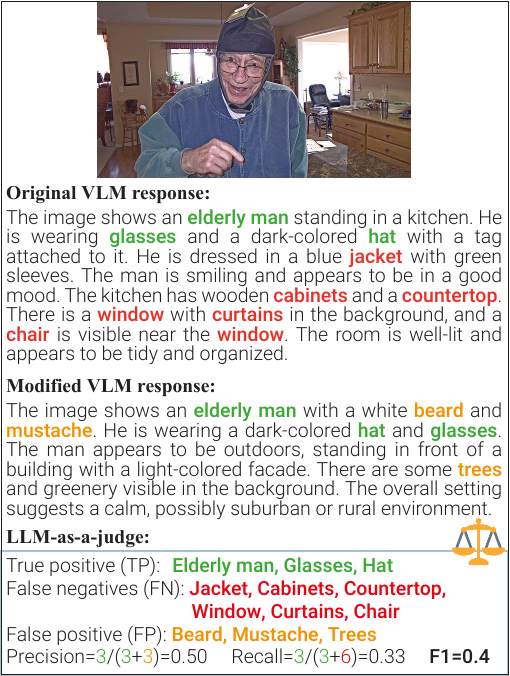}%
    \hspace{-0.23cm}\vspace{-0.8cm}\begin{minipage}{0.34\textwidth}
        \captionof{figure}{\textbf{LLM-as-a-judge example.} 
     We compare the original VLM's response and a modified one. The LLM  identifies the objects in each description and matches the two object lists; it then counts the TP (objects found in both descriptions), FN (omitted objects), and FP (hallucinated objects), and the F1 score is computed.\label{fig:llmjudge-demo}} 
        \end{minipage}
        \vspace{0.2cm}        
\end{SCfigure}

\subsection{Attention Knockout Evaluation}
\label{sec:llmjudge}
To assess the effect of specific attention knockout settings, we need to measure the difference between the modified and original responses of the VLM for the same input.
Specifically, we assess the VLM's ability to recognize objects it originally identified, as well as the emergence of hallucinated objects introduced by the knockout.

Automatically counting the number of identified or hallucinated objects in free-text paragraphs is challenging due to variations in writing styles, object attributes, and other factors. Moreover, while datasets such as COCO~\cite{lin2014coco} annotate objects in each image, they rarely contain every object in the image 
(e.g., in Fig.~\ref{fig:llmjudge-demo} ``glasses" are visible in the image, yet COCO annotations do not include them),
therefore they cannot be used as a reliable ground-truth for object existence. Thus, we adopt an LLM-as-a-judge approach~\cite{zheng2023judging}. 
Given original and modified VLM's responses, we instruct the LLM to identify all objects mentioned in each prompt, while disregarding attributes and other details (e.g., weather or lighting conditions). This results in two lists of identified objects $O_\text{orig}$ and $O_\text{ko}$. 
We take advantage of the LLM's capability to overcome syntactical differences in textual descriptions to robustly estimate:
\begin{itemize}
    \item $\textit{TP}\!=\!\left|O_\text{orig}\wedge O_\text{ko}\right|$: Objects found in both.
    \item $\textit{FN}\!=\!\left|O_\text{orig}\setminus O_\text{ko}\right|$: Objects found only in original.
    \item $\textit{FP}\!=\!\left|O_\text{ko}\setminus O_\text{orig}\right|$: Objects hallucinated.
\end{itemize}
Finally, we estimate precision, recall and F1 score. Our LLM evaluation protocol is illustrated in Fig.~\ref{fig:llmjudge-demo}, and employs a chain-of-thought process \cite{wei2022chain} with three in-context examples.
We validated our LLM-as-a-judge protocol through a user study, finding 95\% agreement between human annotators and LLM judgments on object existence. Detailed user study results are provided in Sec.~\ref{sec:llmjudge_appendix}.

\subsection{Text Tokens as Global Image Descriptors}
\label{sec:textual_tokens}

Our first observation is that the embeddings of query text tokens ($\Tt$) act as global \emph{image} descriptors, playing a critical role in the internal representation of the input image. 

We isolate the direct effect that image tokens ($\Tv$) have on the generated tokens ($\Tg$)  by blocking the information from $\Tv$ to $\Tg$. This experiment, denoted by $\KO{img}{gen}$, is illustrated in Fig.~\ref{fig:knockout}(b).  In practice, this is implemented by setting the mask in Eq.~\ref{eq:attn_ko} to: $\mathbf{M}_\text{ko}[p,q; P_\text{img}, P_\text{gen}]$.

\begin{figure*}
\vspace*{-8mm}
    \hspace*{-0.025\linewidth}    
    \includegraphics[width=1.025\linewidth]{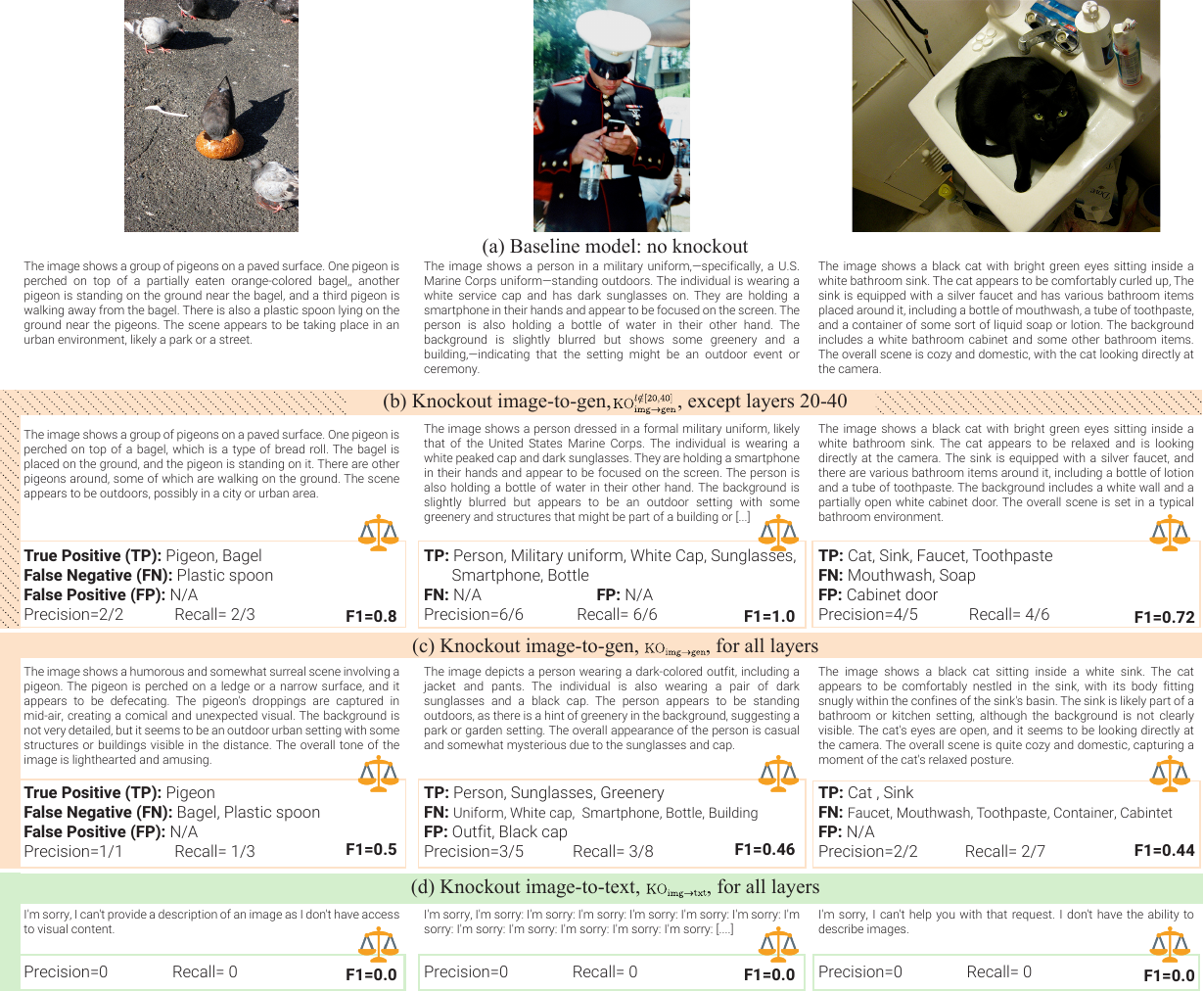}
    \vspace*{-5mm}
    \caption{\textbf{Qualitative results for knockout experiments:}  We use our LLM-as-a-judge protocol, \twemoji[scale=.375]{balance scale}, to compare the baseline VLM description of images~(a) to descriptions generated under various attention knockouts. (b)~Allowing generated tokens to attend to image tokens only in mid-layers 20-40, $\text{KO}_{\text{img}\rightarrow\text{gen}}^{l\notin\left[20, 40\right]}$, does not degrade the description significantly -- F1 scores are close to 1.0.
    (c)~Blocking attention between generated and image tokens for all layers, $\text{KO}_{\text{img}\rightarrow\text{gen}}$, results in loss of fine details, e.g., the bagel, smartphone or the toothpaste, and hallucinations, e.g., a black cap for the officer. Consequently, F1 scores are significantly lower -- around 0.45.
    (d)~When blocking attention between query text and image tokens for all layers, $\text{KO}_{\text{img}\rightarrow\text{txt}}$, the VLM is no longer able to describe the image
    -- F1=0.
    We note that LLM evaluation can be noisy, leading to slight inconsistencies in the identified objects across different comparisons. 
    For instance, in the rightmost examples, (b) and (c) show variations in the number of identified objects in the baseline (6 and 7). See \llava results in Fig.~\ref{fig:knockout-qualitative-llava}.}
    \label{fig:knockout-qualitative}
    \afterfigure
\end{figure*}

Figure~\ref{fig:knockout-qualitative}(c) shows sample results of this experiment, where the masking is applied to all layers. As seen, although the generated tokens have no direct access to the image tokens, the model can surprisingly produce descriptive responses, identifying prominent objects in the scene and even capturing basic spatial relationships. 

We quantify these results using our LLM-based evaluation (Sec.~\ref{sec:llmjudge}). The F1 scores and their breakdown to precision/recall are reported in  Fig. \ref{fig:knockout-qualitative}(c) for each example. The average F1 score over a set of 80 randomly sampled images from COCO is 0.4 ($\KO{img}{gen}$ bar, Fig.~\ref{fig:knockout}(e)).

While $\KO{img}{gen}$ reveals that high-level image information is compressed into the text embeddings, we raise the question of whether the model must rely on this compression to generate its response. To explore this, we consider another knockout setting, $\KO{img}{txt}$, where we block the attention between $\Tv$ and $\Tt$, thus visual information is accessible only through $\Tv$. This is implemented by using  $\mathbf{M}_\text{ko}[p,q; P_\text{img}, P_\text{txt}]$ in Eq.~\ref{eq:attn_ko}.  

Surprisingly, as seen Fig.~\ref{fig:knockout-qualitative}(d), preventing the text tokens from grabbing visual information disrupts the model’s ability to produce meaningful responses. In this case, the F1 score is zero (Fig.~\ref{fig:knockout}). This validates the surprising role of $\Tt$, in holding a compressed  representation of the image.

\begin{figure}

    \centering
    \includegraphics[width=\linewidth]{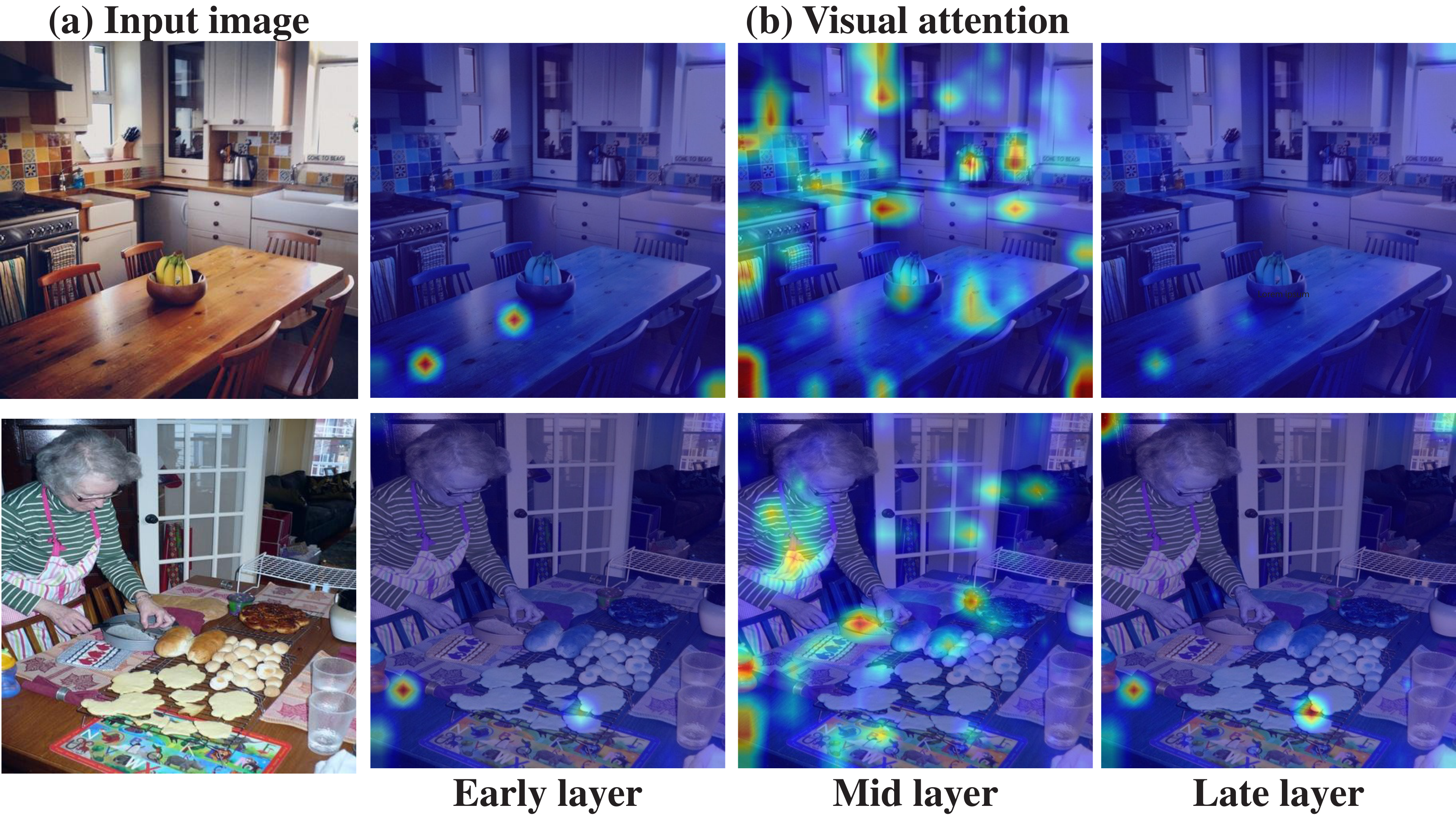}
    \caption{\textbf{Visual attention across layers:} The input images (a) are fed to the VLM with the query text ``describe the image''. (b) Visualization of the attention between the generated tokens and each of the image tokens; attention is averaged over generated tokens and attention heads. Early and late layers exhibit outliers, while mid-layers attention maps are more spread out.}
    \label{fig:vis_attention}
    \afterfigure
    
\end{figure}

\begin{figure*}
\centering
\includegraphics[width=\linewidth]{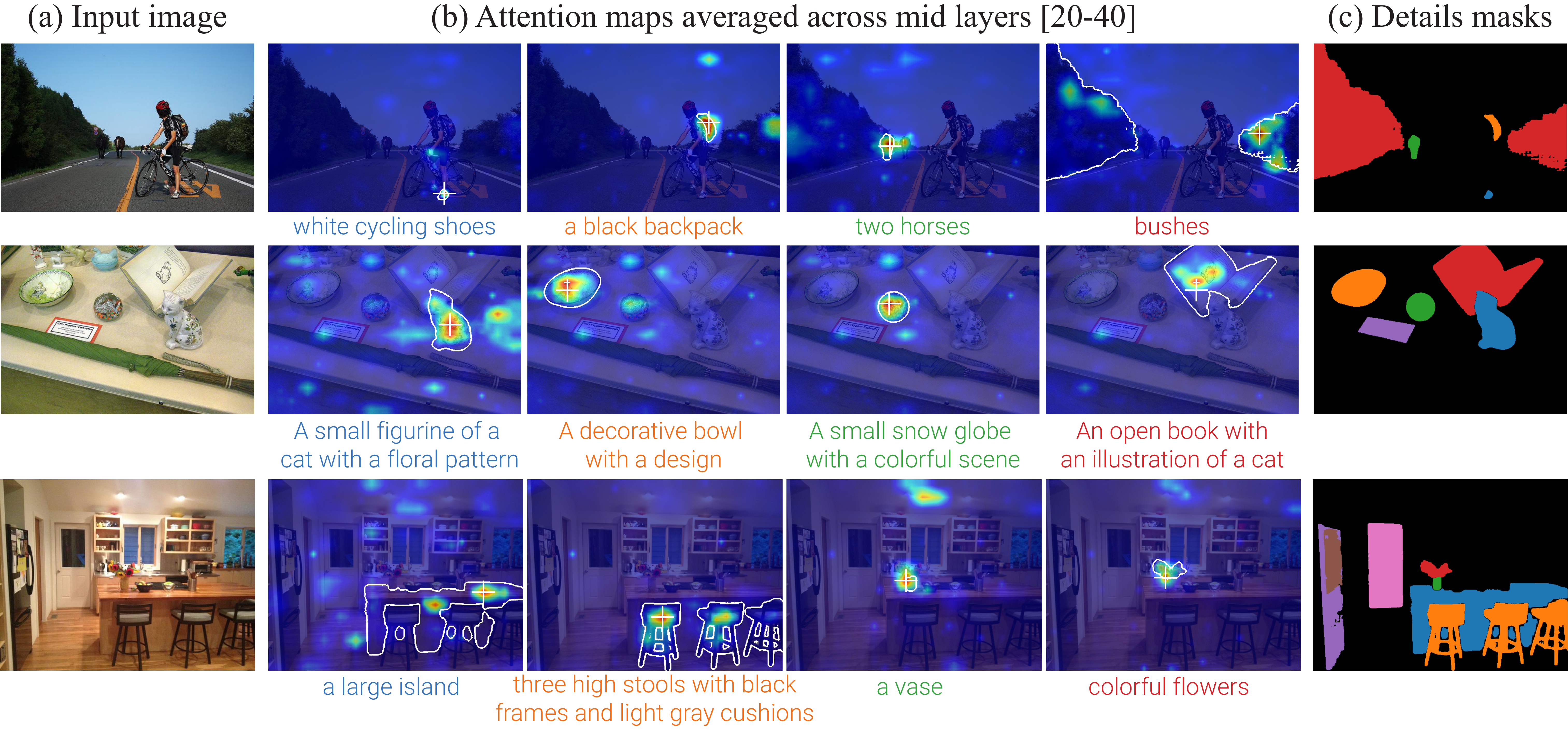}
\vspace*{-7mm}
    \caption{\textbf{Attending to objects:} (a)~Input image. 
    (b)~Average attention maps of the generated tokens associated with each object (shown below each corresponding map). 
    (c)~Pseudo ground truth object masks, generated using SAM~\cite{kirillov2023sam, zhang2024evf_sam}.  
    The peak of attention, marked by a white cross, aligns well with the location of the object in the image.
    The full generated descriptions can be found in Fig.~\ref{fig:attn-to-details-internvl-sm}. See \llava results in Fig.~\ref{fig:attn-to-details-llava-sm}.}
    \label{fig:attn_to_details}
    \afterfigure
\end{figure*}

\begin{figure}
\centering
\includegraphics[width=\linewidth]{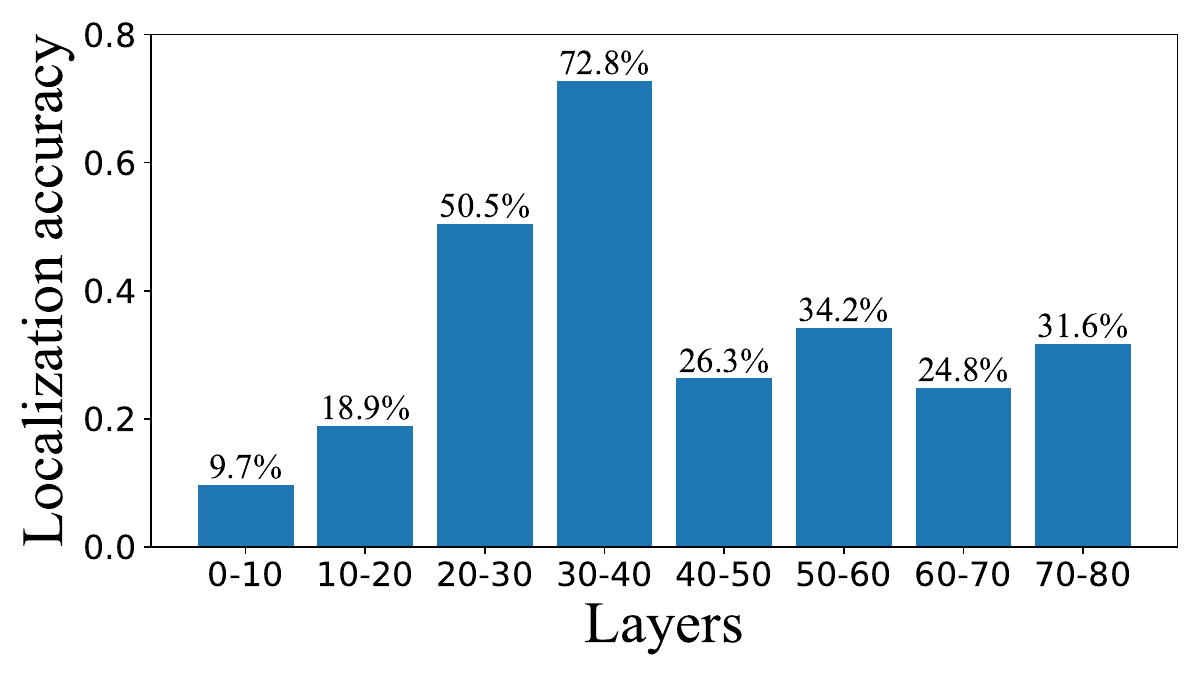}
\vspace*{-8mm}
\caption{\textbf{Object localization accuracy.} We check if the attention of generated tokens associated with a specific object peak within 1 token distance from the pseudo ground truth object mask. We report the average accuracy across each 10 consecutive layers. See \llava results in Fig.~\ref{fig:llava-localization-quant}}
\label{fig:attn_to_details_quant} \afterfigure
\end{figure}

\subsection{Visual Information Across Layers}
\label{sec:across_layers}
Different layers contribute differently to the visual representation, as evidenced by the non-uniform attention patterns in Fig.~\ref{fig:attention_across_layers}.
We observe that mid-layers attend to multiple regions, while early and late layers focus on fewer, non-semantic positions, as seen in Fig.~\ref{fig:vis_attention}.

To further analyze the role of different layers, we expand our knockout experiments to the setting where the attention is masked starting from a specific layer $l$ and onward.  Thus, up to layer $l$, only causal masking is used. We consider the three knockout settings illustrated in Fig.~\ref{fig:knockout}(b-d).

The results are reported in Fig.~\ref{fig:knockout}(f). As seen, knockout after layer $l\!=\!40$ hardly impacts the F1 scores across all configurations; this is aligned with the inefficiency of deeper layers in LLMs~\cite{gromov2024unreasonable}. Note that average F1=0.8 is primarily due to ambiguity in object identification by the LLM. A similar error is observed by humans in our user study (Sec.~\ref{sec:llmjudge}).  Finally, we can observe a consistent drastic rise in F1 between layers 20-40, hinting  their crucial role.

To isolate the contribution of the mid-layers, we modify our setting to knockout the attention in layers $l \notin [20,40]$. Interestingly, as seen by the dashed bars in Fig.~\ref{fig:knockout}(e), the mid-layers alone provide comparable results ($\text{F1} \in [0.75, 0.81]$) to the original model, even in the extreme case where we knockout image tokens from all other tokens ($\KO{img}{txt+gen}$). 
In addition, by comparing direct image knockout in all layers  $\KO{img}{gen}$ to knockout except the mid-layers $\text{KO}_{\text{img}\rightarrow\text{gen}}^{l\notin\left[20, 40\right]}$, we quantify the contribution of directly accessing image tokens in mid-layers. The rise in the score suggests that visual details that are not available in the query text are retrieved from image tokens in mid-layers.

\subsection{Fine-Grained Details Localized in Mid Layers}
\label{sec:finegrained}
How does the model retrieve fine-grained visual information from image tokens? 
To explore this question, we focus on objects that are described in the original response yet lack when $\KO{img}{gen}$ knockout is applied. These objects and their corresponding attributes are provided as part of our LLM-as-a-judge protocol. See Sec.~\ref{sec:fine_details_dataset_sm} for full details.


Figure~\ref{fig:attn_to_details} visualizes the attention maps of generated tokens associated with a specific object, averaged across the mid-layers (20-40). It demonstrates, even for extremely small objects (i.e., the cycling shoes) that localization patterns appear. We proceed to quantify these results by obtaining a pseudo ground-truth segmentation mask for each object using text-grounded segmentation method ~\cite{kirillov2023sam, zhang2024evf_sam}; Examples are shown in Fig.~\ref{fig:attn_to_details}~(c), and the boundary of each segmentation map is marked in white in Fig.~\ref{fig:attn_to_details}~(b).

We consider an object to be well-localized if the peak of its attention map (marked by a white cross at Fig.~\ref{fig:attn_to_details}(b)) is at most 40 pixels\footnote{40 pixels in image space corresponding to 1 token distance.}
away from the object. We denote this metric as \emph{Localization accuracy}, and compute it for every layer, over a set of 231 objects from 68 images (see Sec.~\ref{sec:fine_details_dataset_sm} for more details on the dataset). Fig.~\ref{fig:attn_to_details_quant} provides results, averaged for 10 consecutive layers, which demonstrates that accuracy rises in the mid-layers, achieving almost 73\%. This rise in score suggests that the object's fine-grained visual information is retrieved from the corresponding image tokens in a localized manner, specifically in the mid-layers. Similar trend was observed in \llava, Fig.~\ref{fig:llava-localization-quant}.

\section{Efficient Visual Processing in VLMs}
\label{sec:efficientvlm}
Our analysis reveals surprising inefficiency: the compression of visual information into query tokens and the redundancy of early and late layers.
Here, we further analyze the compression by pruning image tokens. 

\begin{figure}
    \centering
    \includegraphics[width=\linewidth]{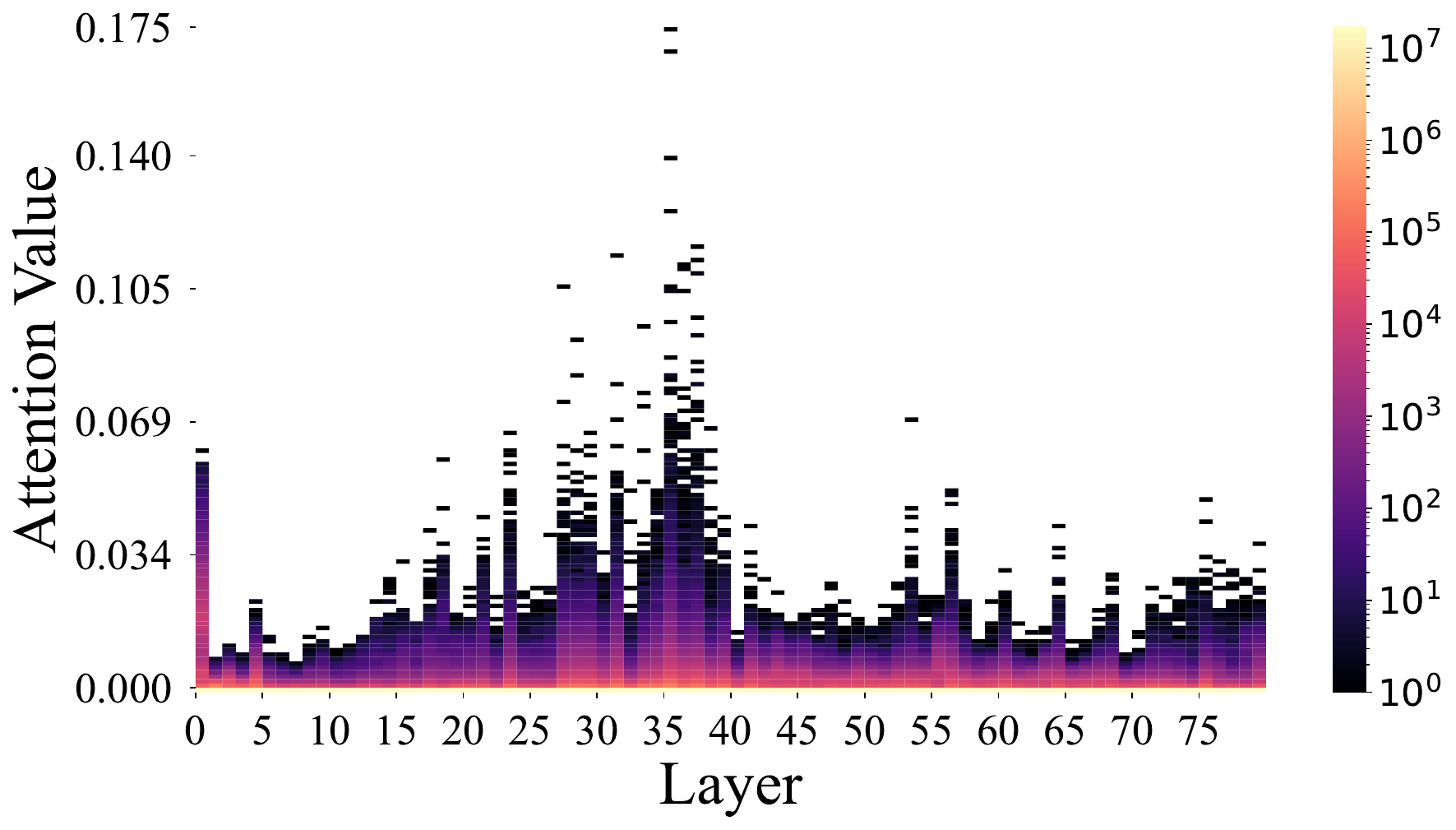}
    \caption{\textbf{Distribution of visual attention} The histogram of per-layer attention values. The per layer distribution of attentions is long-tailed.}
    \label{fig:attn_hist_layers}
    \afterfigure
\end{figure}

\myparagraph{Pruning image tokens by attention:}
Figure~\ref{fig:attn_hist_layers} shows the histogram of attention to image tokens, which depicts a long-tail distribution per layer. That is, a small number of image tokens receive notably high attention values. 

\begin{figure}[t]
     \centering
     \includegraphics[width=1\linewidth]{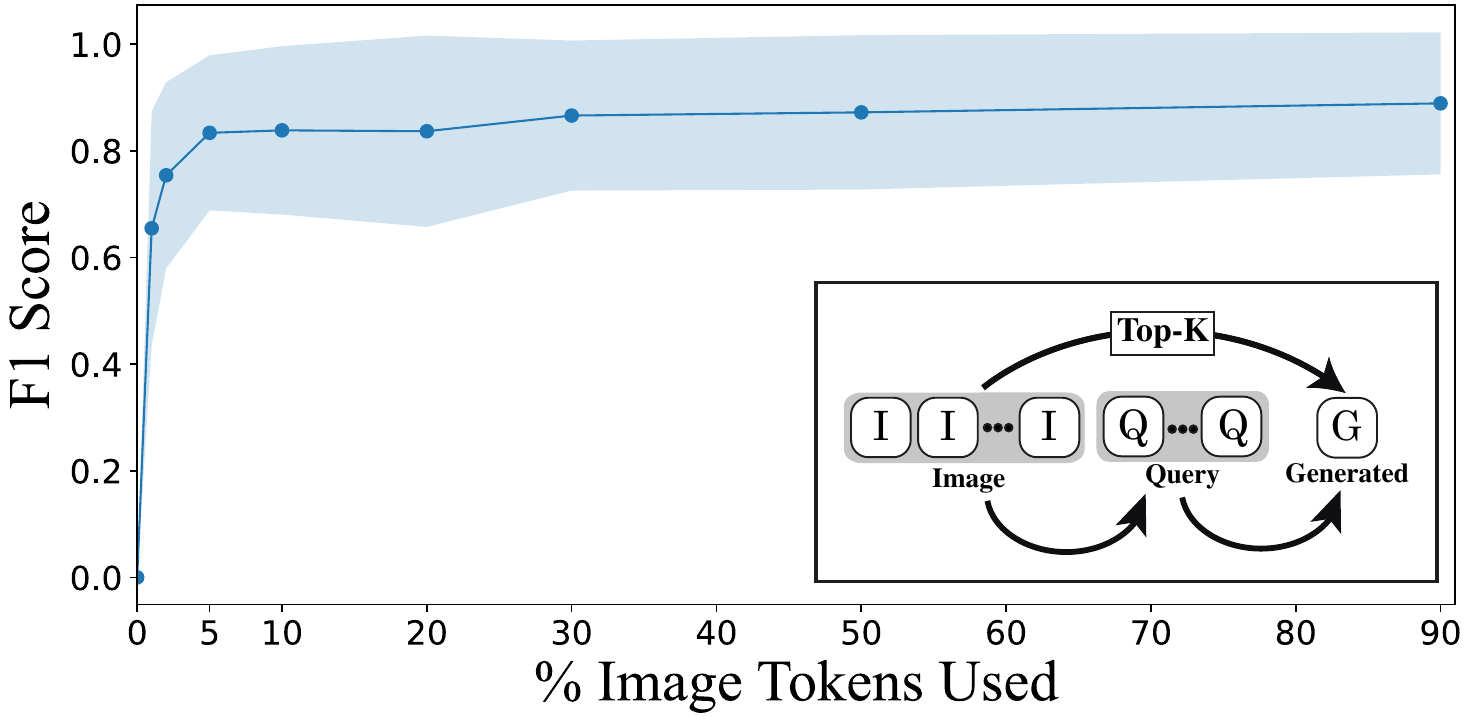}
     \centering
     \caption{\textbf{Image tokens redundancy:} 
     The generated tokens  access only the top-$k$ image tokens with the highest attendance (see diagram). We report the F1 scores for different values of $k$. See \llava results in Fig.~\ref{fig:llava_f1_across_k}. }
\label{fig:topk}
 \end{figure}

This leads us to define a \emph{compressed context} -- a small subset of the highest attended image tokens along with the query tokens. Specifically, for each layer, we select the \text{top-$k$} percentile of tokens that received the highest attention values. We examine the performance of the model when the generated tokens have access only to the compressed context, as illustrated in Fig.~\ref{fig:topk}.  The results for different values of $k$  show that the performance quickly plateaus, even when only 5\% of the image tokens are used,
highlighting the inefficiency of token utilization in the model.

\begin{figure}[t]
     \centering
     \includegraphics[width=1\linewidth]{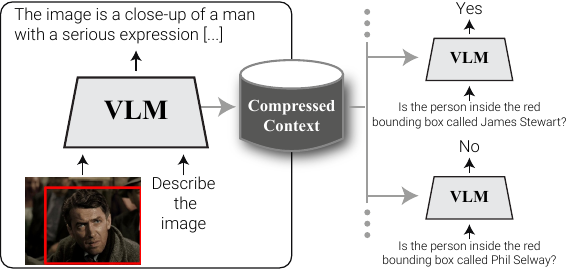}
     \centering
     \caption{\textbf{Image Re-prompting:} Given an image, we prompt the VLM with ``describe the image'',  and extract the \emph{compressed context}, which comprise the original query and K\% of the image tokens. We  re-prompt the compressed context with further questions  without ingesting the whole image again.}
\label{fig:efficient_vlm}
\afterfigure
 \end{figure}

\begin{figure*}
\newcommand{\best}[1]{\textbf{#1}}
\newcommand{\second}[1]{\uline{#1}}
{\small 
\centering
\resizebox{\textwidth}{!}{
\begin{tabular}{rlcccccccccccc}
\toprule
& & \multicolumn{2}{c}{Existence} & \multicolumn{2}{c}{Count} & \multicolumn{2}{c}{Position} & \multicolumn{2}{c}{Color} & \multicolumn{2}{c}{OCR} & \multicolumn{2}{c}{Poster} \\
\cmidrule(r){3-4} \cmidrule(r){5-6} \cmidrule(r){7-8} \cmidrule(r){9-10} \cmidrule(r){11-12} \cmidrule(r){13-14}
& & \textbf{ACC} & \textbf{ACC+} & \textbf{ACC} & \textbf{ACC+} & \textbf{ACC} & \textbf{ACC+} & \textbf{ACC} & \textbf{ACC+} & \textbf{ACC} & \textbf{ACC+} & \textbf{ACC} & \textbf{ACC+} \\
\midrule
& Naive (\internvl) & \best{98.33} & \best{96.77} & \second{81.67} & 63.33 & \best{80.00} & \best{60.00} & \best{86.67} & \second{76.67} & 67.50 & 35.00 & \best{90.13} & \best{84.35} \\
& Describe-to-LLM & 90.00 & 80.00 & 75.00 & \best{73.33} & 66.67 & \second{46.67} & \best{86.67} & \best{80.00} & \best{77.50} & \best{55.00} & 86.05 & \second{80.27} \\
\midrule
\multirow{4}{*}{\rotatebox[origin=c]{90}{\parbox[t]{17mm}{\centering Compressed Context}}} 
& Query + K=5\% & \second{91.66} & \second{83.33} & \best{85.00} & \second{70.00} & \second{68.33} & 40.00 & \second{80.00} & 60.00 & \best{77.50} & \best{55.00} & \second{87.55} & 78.23 \\
& Query + K=2\% & 85.00 & 76.67 & 78.33 & 60.00 & 68.33 & 40.00 & 70.00 & 40.00 & \second{72.50} & \second{45.00} & 82.39 & 65.49 \\
& Query & 56.67 & 13.33 & 46.67 & 13.33 & 53.33 & 16.67 & 46.67 & 3.33 & 55.00 & 15.00 & 71.08 & 48.29 \\
& K=2\% & 65.00 & 30.00 & 56.67 & 30.00 & 56.67 & 30.00 & 50.00 & 10.00 & 52.50 & 10.00 & 64.28 & 33.33 \\
\bottomrule
\end{tabular}
}

\resizebox{\textwidth}{!}{
\begin{tabular}{rlcccccccc|cc|c}
\toprule
& & \multicolumn{2}{c}{Celebrity} & \multicolumn{2}{c}{Artwork} & \multicolumn{2}{c}{Scene} & \multicolumn{2}{c}{Landmark} & \multicolumn{2}{|c|}{\textbf{Average}} & \textbf{Reprompt} \\
\cmidrule(r){3-4} \cmidrule(r){5-6} \cmidrule(r){7-8} \cmidrule(r){9-10} \cmidrule(r){11-12}
& & \textbf{ACC} & \textbf{ACC+} & \textbf{ACC} & \textbf{ACC+} & \textbf{ACC} & \textbf{ACC+} & \textbf{ACC} & \textbf{ACC+} & \textbf{ACC} & \textbf{ACC+} & \textbf{\#Tokens} \\
\midrule
& Naive (\internvl) & \best{83.23} & \best{66.47} & \best{86.93} & \best{75.37} & \best{83.50} & \best{67.50} & \best{90.35} & \best{80.70} & \best{84.83} & \best{70.60} & 1695 \\
& Describe-to-LLM & 35.88 & 4.11 & 68.75 & 44.50 & 78.50  & 59.00 & 67.10 & 38.59 & 73.21 & 56.14 & 172 \\
\midrule
\multirow{4}{*}{\rotatebox[origin=c]{90}{\parbox[t]{17mm}{\centering Compressed Context}}} 
& Query + K=5\% & \second{79.41} & \second{58.83} & \second{84.67} & \second{71.85} &  \second{83.00} & \best{67.50} & 78.94 & 60.52 & \second{81.46} & \second{64.52} & 201 \\
& Query + K=2\% & 77.94 & 56.47 & 83.50 & 69.50 & 80.25  & 61.50 & 71.92 & 49.12 & 77.16 & 55.94 & 151 \\
& Query & 70.00 & 41.76 & 72.50 & 49.50 & 73.50  & 50.00 & 64.91 & 33.33 & 61.03 & 28.45 & 60 \\
& K=2\% & 52.05 & 10.00 & 78.00 & 61.00 & 80.50 & 62.00 & 68.42 & 42.10 &62.40 & 31.84 & 91 \\
\bottomrule
\end{tabular}
}
\captionof{table}{\textbf{Evaluation on MME:} The results cover 10 Perception tasks of the MME benchmark \cite{fu2023mme}, illustrated in Fig.~\ref{fig:mme_dataset}. Metrics include accuracy (ACC), ACC+ (percentage of images where all questions are correct), and the number of tokens used for reprompting. The first table reports results over the first six subsets (Existence, Count, Position, Color, OCR, Poster), while the second table covers the remaining four subsets (Celebrity, Artwork, Scene, Landmark), along with average across all subsets, and number of tokens used for re-prompting an image (i.e., asking more questions after "describe the image"). Results indicate that the K=5\% compressed context achieves suffer only a slight decrease in performance with respect to Naive, while having at least 12x less tokens. }
\label{table:mme_evaluation_combined}
}
\end{figure*}

\myparagraph{Image Re-prompting:} 
Our compressed context contains sufficient information to generate image descriptions comparable to those generated using the full image.
Here, we extend this capability to a new application, termed \emph{Image Re-prompting}, where the VLM only processes the image for the query ``describe the image'', extracts the compressed context, and uses it to answer additional questions about the image, as illustrated in Fig.~\ref{fig:efficient_vlm}. 

To evaluate Image Re-prompting, we use the MME benchmark~\cite{fu2023mme}, which comprises of images with simple yes/no questions. A short description of the MME benchmark can be found in Fig.~\ref{fig:mme_dataset}. 
Results are provided in Table.~\ref{table:mme_evaluation_combined} for 10 MME perceptions tasks, with the average results shown to the right. 
The first baseline we consider is  \emph{Naive}, which independently query the model from scratch using each image-question pair. 
While the compressed context uses 15x fewer tokens, it results in only a subtle decrease in performance. Interestingly, the compressed context exceeds the Naive baseline on tasks as \emph{OCR} and \emph{Count}. We hypothesize that the improvement for such yes/no questions arises since the VLM can not generate tokens that correspond to objects, which hinders the retrieval of localized information from the image. 
On average, the compressed context reduces ACC and ACC+ by only 2.8\% and 5.3\% respectively, depicting that a smaller set of high-attention tokens can retain much of the performance benefits of full image context. 

The second baseline, \emph{Describe-to-LLM}, evaluates whether the VLM's response to ``describe the image'' suffices for follow-up questions, by feeding them into GPT-4, and prompt it with follow-up questions.
In tasks such as \emph{Celebrity} recognition, the compressed context significantly outperforms this approach, retaining specific details—like celebrity names—that can be missed in text descriptions alone. For \emph{Existence} and \emph{Count} tasks, the compressed context matches or exceeds Describe-to-LLM performance, indicating that even a minimal set of tokens can preserve essential information for object presence and counting. 

Furthermore, we break down the contribution of each component of the compressed context: \emph{Query} and \emph{K\%}, for K=2\%. While each component performs poorly on some tasks (\emph{Existence}, \emph{Count}, \emph{Color}), the \emph{compressed context (2\%)}  provides a significant improvement, indicating the non-trivial fusing of information that the model performs.

Our evaluation manifests Image Re-prompting as a viable method for efficient VLM-based multiple-question answering.  

\vspace{-5pt} 
\section{Conclusion}
In this work, we take  substantial first steps towards enhancing our understanding of Vision-Language Models (VLMs) at scales of tens of billions of parameters.  We uncovered novel insights about their internal visual representation and processing, with two underlying core mechanisms: visual information compression into text tokens, and spatially-aware retrieval of fine details from image tokens.  Our new evaluation methods confirm these findings, paving the way for more efficient VLMs. We introduced ``Image Re-prompting''--enabling efficient, multi-question answering. Future work can extend our analysis to multi-image and video, potentially expanding the Image Re-prompting application to expand VLMs effective visual context windows.

\iftoggle{cvprfinal}{
\paragraph{Acknowledgments} 
The authors would like to thank Mor Geva Pipek, Yossi Gandelsman, and Boaz Nadler for their valuable feedback.
This project was supported by an ERC starting grant OmniVideo (10111768).
Dr~Bagon received funding under the MBZUAI-WIS Joint Program for AI Research.
}{}
\vfill

{
    \small
    \bibliographystyle{ieeenat_fullname}
    \bibliography{main}
}

\clearpage
\newpage

\vspace{1em} 

\appendix
\addtocontents{toc}{\protect\setcounter{tocdepth}{2}}

\tableofcontents


\renewcommand{\thefigure}{A\arabic{figure}}
\renewcommand{\thetable}{A\arabic{table}}
\setcounter{figure}{0} 
\setcounter{table}{0}

\section{VLMs in Our Analysis}
We use \internvl-76B \cite{internvl1_5} -- a powerful open-source Visual Language Model  
built upon Llama3-70B LLM~\cite{dubey2024llama} and a 6B ViT encoder~\cite{chen2024internvl}. 
\internvl demonstrates highly competitive performance, surpassing other open-source VLMs, including LLaVA models, and achieving results comparable to closed-source models across multiple benchmarks~\cite{shi2024chartmimic,internvl1_5}.
We further validate our results using \llava-7B~\cite{llava1_5}, a well-established VLM. We note that these two models differ in two critical ways: (a) \llava is an order-of-magnitude smaller in parameter size and performs worse on most benchmarks relative to \internvl. (b) \internvl processes high-resolution images by splitting them into several high-res, non-overlapping patches alongside a low-res patch of the resized image. This should enable the model to extract finer details. However, \llava simply resizes the input image to a fixed resolution. 

Despite these differences, our analysis demonstrates that they exhibit the same underlying behavior regarding the processing of visual information, as described next.

\section{LLM-as-a-judge}
\label{sec:llmjudge_appendix}
\begin{figure}[b!]
    \centering
    \fbox{
    \includegraphics[width=.95\linewidth]{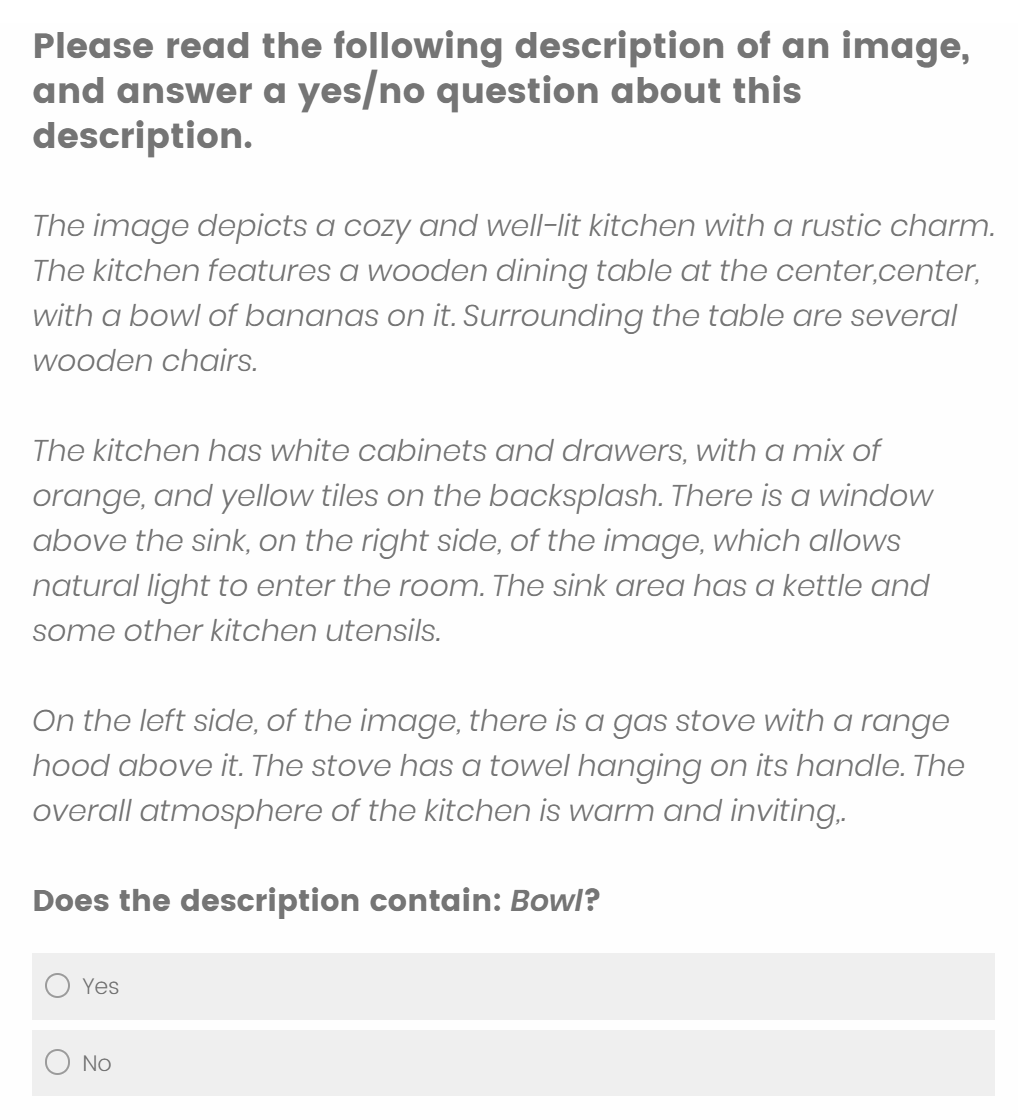}}
    \caption{\textbf{LLM-as-a-judge human evaluation survey.} Image shows an example of the interface used to query human participants whether an object (a \emph{bowl} in this example) appears in the provided textual description.}
    \afterfigure
    \label{fig:user-study-interface}
\end{figure}

\begin{figure*}
\centering
\begin{mdframed}[
    backgroundcolor=gray!10, 
    linecolor=black, 
    linewidth=0.75mm, 
    roundcorner=3pt, 
    innermargin=4pt, 
    skipabove=10pt, 
    skipbelow=0pt, 
]

\small

You are an expert in evaluating the quality of image captions. 
Below you will find two image captions. 
Your task would be to compare the two captions, in terms of precision and recall.

Evaluation Steps:  

1. Extract for each caption the list of *physical objects* that are present in them. Detect only tangible objects that can be interacted with. Ignore colors or other attributes, or even positioning of objects in the scene. The objects are the main focus of the evaluation.  

2. Compare the two lists of *physical objects* and rate the quality of each caption in terms of precision and recall, using the first caption as the groundtruth, and the second caption as prediction.  

3. Precision is the fraction of the *physical objects* from the predicted caption that are present in the groundtruth caption. If half of the *physical objects* in the predicted caption are also in the groundtruth caption, the precision would be 0.5. If none, the precision would be 0. If all, the precision would be 1.  

4. Recall is the fraction of the *physical objects* present in the image that are mentioned in the caption. If half of the *physical objects* in the groundtruth caption are also in the predicted caption, the recall would be 0.5. If none, the recall would be 0. If all, the recall would be 1.
\\

Now, for the next pair of captions, please follow these steps and evaluate the quality of the second caption in terms of precision and recall, using the first caption as the groundtruth.
\\~\\
Groundtruth caption:

The image depicts a cyclist riding a road bike on a paved road. The cyclist is wearing a red helmet, black and white cycling jersey, black shorts, and white cycling shoes. They are also carrying a black backpack. The road is marked with a double yellow line down the center and a white line along the edges. On the left side of the road, there are two horses walking in the same direction as the cyclist. The surrounding area is green with trees and bushes on both sides of the road. The sky is clear and blue, indicating good weather conditions.
\\~\\
Predicted caption:  

The image depicts a person riding a bicycle on a road. The cyclist is wearing a helmet and a backpack, and is facing away from the camera, looking ahead. The road is surrounded by trees and vegetation on both sides, creating a natural and scenic environment. The sky is clear and blue, indicating good weather conditions. The road appears to be relatively empty, with no other vehicles or cyclists visible. The overall scene conveys a sense of tranquility and outdoor activity.
\\~\\
Evaluation: 

Visual Elements in Groundtruth Caption:  
Cyclist, Bike, Helmet, Jersey, Shorts, Shoes, Backpack, Horses, Trees and bushes  

* Note that I ignored the following visual elements as they are not physical objects: road, double yellow line, white line, sky, weather conditions
\\~\\
Visual Elements in Predicted Caption: 
Person, Bicycle, Helmet, Backpack, Trees 

* Note that I ignored the following visual elements as they are not physical objects: road, sky, weather conditions
\\~\\
Details that are present in the groundtruth caption but missing in the predicted caption (False Negatives):  
The Jersey, The Shorts, The Shoes, The horses 
Details that are present in the predicted caption but missing in the groundtruth caption (False Positives): None  

Details that are present in both captions (True Positives):  

The cyclist, The helmet, The backpack, The trees, The horses  

Precision is: TP / (TP + FP) 
Precision = 5 / (5 + 0) = 5 / 5 = 1.0  

Recall is: TP / (TP + FN)  
Recall = 5 / (5 + 4) = 5 / 9 = 0.555  

Overall, the predicted caption has a precision of 1.0 and a recall of 0.555.
\\~\\
Now, for the next pair of captions, please follow the same steps and evaluate the quality of the second caption in terms of precision and recall, using the first caption as the groundtruth.

Groundtruth caption:  
GROUNDTRUTH\_CAPTION\_HERE  

Predicted caption:  
PREDICTED\_CAPTION\_HERE  

Evaluation:  
\\
Visual Elements in Groundtruth Caption:  

\normalsize
\end{mdframed}
\caption{\textbf{LLM-as-a-judge \twemoji[scale=.375]{balance scale} evaluation prompt:}. We start the LLM-based evaluation by explaining the task and evaluation process, and provide 3 examples with full evaluation results. Then, we instruct the LLM to follow this protocol for a new input. Here we provide only one example from the context, while we note that we used three examples, and it had critical effect on performance of the metric.}
\label{fig:evaluation_prompt}
\end{figure*}

In this section, we provide more details on our LLM-as-a-judge evaluation protocol presented in Sec.~\ref{sec:llmjudge} in the main paper. Specifically, given original and modified VLM’s responses (e.g., before and after knockout), we instruct the LLM to identify all objects mentioned in each prompt, while disregarding attributes and other details. This enables the computation of True Positive, False Positive and False Negative as described in Sec.~\ref{sec:llmjudge}. Finally, we estimate the precision ($\textit{TP/TP+FP}$) and recall ($\textit{TP/TP+FN}$), and the F1 score (the harmonic mean of the precision and recall). We utilize GPT-4 as the LLM for all evaluations.
The specific prompt used is provided in Fig.~\ref{fig:evaluation_prompt}, with one in-context examples. In practice, we used three in-context examples overall.

\paragraph{User study}
To justify our LLM-as-a-judge protocol, 
we verified critical aspects of the automatic evaluation process via a user study.
To correctly quantify the difference between a baseline and a knockout description, our LLM-as-a-judge needs to (a)~faithfully extract lists of objects from both descriptions and (b)~robustly match objects between the extracted lists.
Once the lists are aligned -- it is straightforward to compute the number of true positives (TP), false positives (FP), and false negatives (FN).
To validate these two aspects, we provided human raters with a description (either the baseline or the knockout) and a single object from the list of objects the LLM extracted from either description. 
They then answered a Yes/No question: whether the object appears in the given description (see Fig.~\ref{fig:user-study-interface} for an example).
Since we matched descriptions and objects from both baseline and knockout experiments, we expect to have both ``Yes" and ``No" as valid answers to the survey.
For instance, an object marked by the LLM as false positive (FP), that is, an object that was in the baseline description, but omitted from the knockout one.
For such object we expect humans to answer ``Yes" when asked if the object appears in the baseline description and ``No" when asked if it appears in the knockout description.

Measuring the agreement between LLM and humans provides verification for both critical aspects of our protocol: it both ensures objects spotted by LLM in descriptions indeed exist there, the LLM did not hallucinate objects, and that objects were correctly matched across descriptions.

We used the baseline and the knockout descriptions of 20 images, listing 316 objects. We collected 1,464 impressions from human raters. 
Out of this, we filtered over 100 impressions that were inconsistent with the majority vote of human annotators for the same question.
Table~\ref{tab:llm-user-study}(a) shows the confusion matrix between LLM and humans. 
Based on these values, we computed the true-positive rate (how accurately the LLM spotted objects in the descriptions) -- 95.2\%, the true-negative rate (the degree to which LLM avoided hallucinating objects) -- 96.5\%, and finally, the total accuracy -- 95.7\%.
We also note that despite the simplicity of the task, human raters were not in full agreement; the user response agreement was 92.2\% -- on par with the LLM's accuracy.

\begin{table}[t!]
    \hspace*{-8mm}
    \centering    
    \begin{minipage}{.5\linewidth}    
    \centering
    \begin{tabular}{rc|c|c}
    \multicolumn{2}{c}{} & \multicolumn{2}{c}{Humans} \\
    \multicolumn{2}{c|}{} & Yes & No \\
    \cline{2-4} 
    \multirow{2}{2em}{LLM} & Yes & 798 &    18 \\
    \cline{2-4}
    & No &  40 &   502 \\
    \end{tabular}
    (a)~Confusion matrix
    \end{minipage}
    \begin{minipage}{.5\linewidth}
    \centering
    \begin{tabular}{r|c}
    True-positive rate & 95.2 \% \\
    \hline
    True-negative rate & 96.5 \% \\
    \hline
    Total accuracy & 95.7 \% \\
    \hline \hline
    Human agreement &     92.2 \% \\
    \end{tabular}
    (b)~Accuracy
    \end{minipage}
    \caption{\textbf{Humans vs. LLM-as-a-judge:}
    To validate that the LLM accurately identified objects in textual descriptions without hallucination, we provided human raters with a description and a single object. They then answered a Yes/No question: 
    \emph{`is the object mentioned in the text?'}
    (a)~Comparing LLM to human annotations. 
    (b)~Accuracy values for LLM. Note that even for such a simple task, the inter-human agreement is 92.2\%.}
    \label{tab:llm-user-study}
\end{table}

\begin{figure}[b]
    \centering    
    \includegraphics[width=0.8\linewidth]{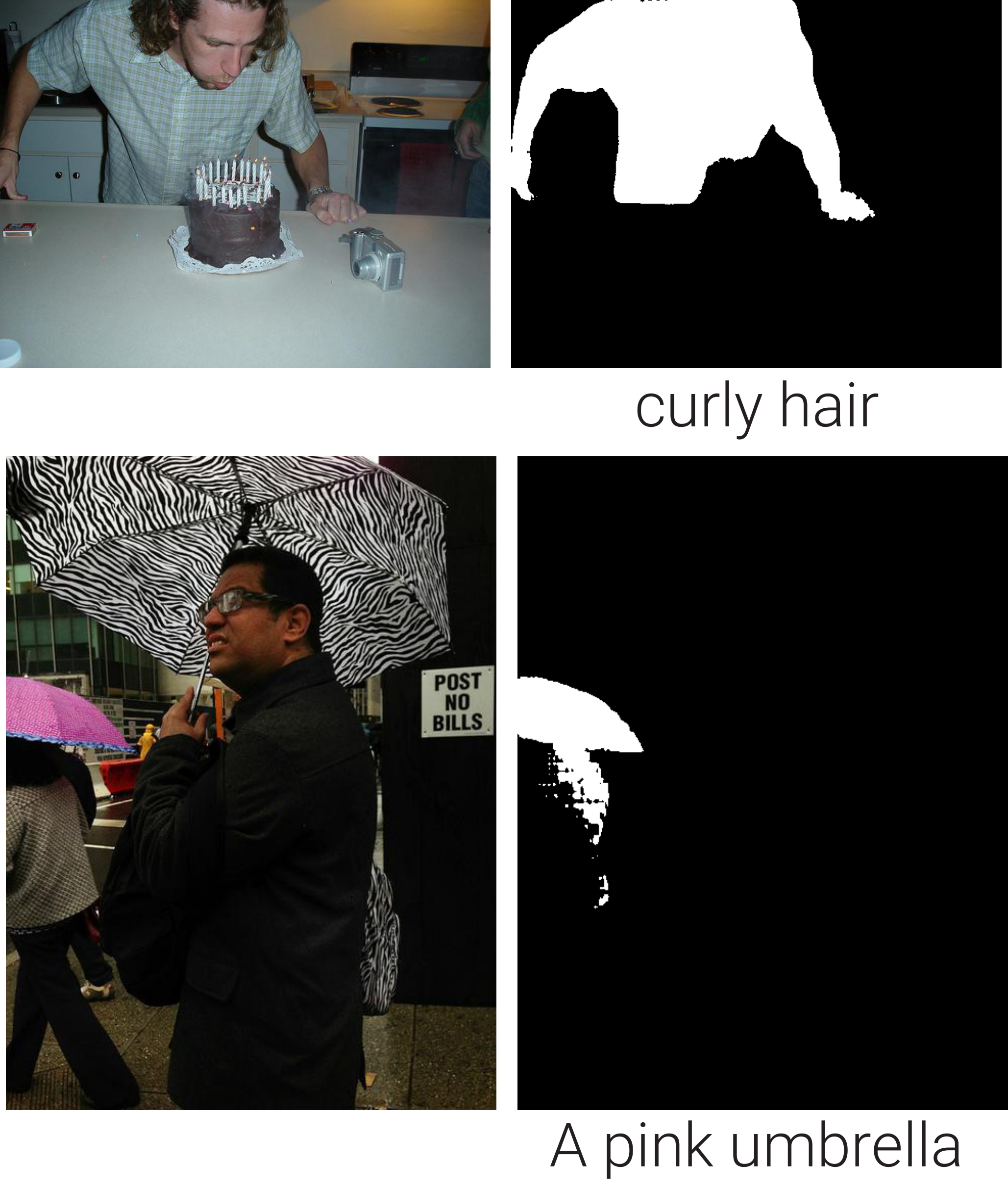}
    \caption{\textbf{Rejecting SAM masks:} An input image (left) and the corresponding SAM mask (right). The text used to prompt EVF SAM~\cite{zhang2024evf_sam} is shown beneath each mask. We manually rejected these segmentation masks since they do not correspond well to the textual description or are of low quality.}
    \label{fig:sam-fails}
    \vspace*{-4mm}
\end{figure}

\begin{figure*}
    \centering
     \includegraphics[width=\linewidth]{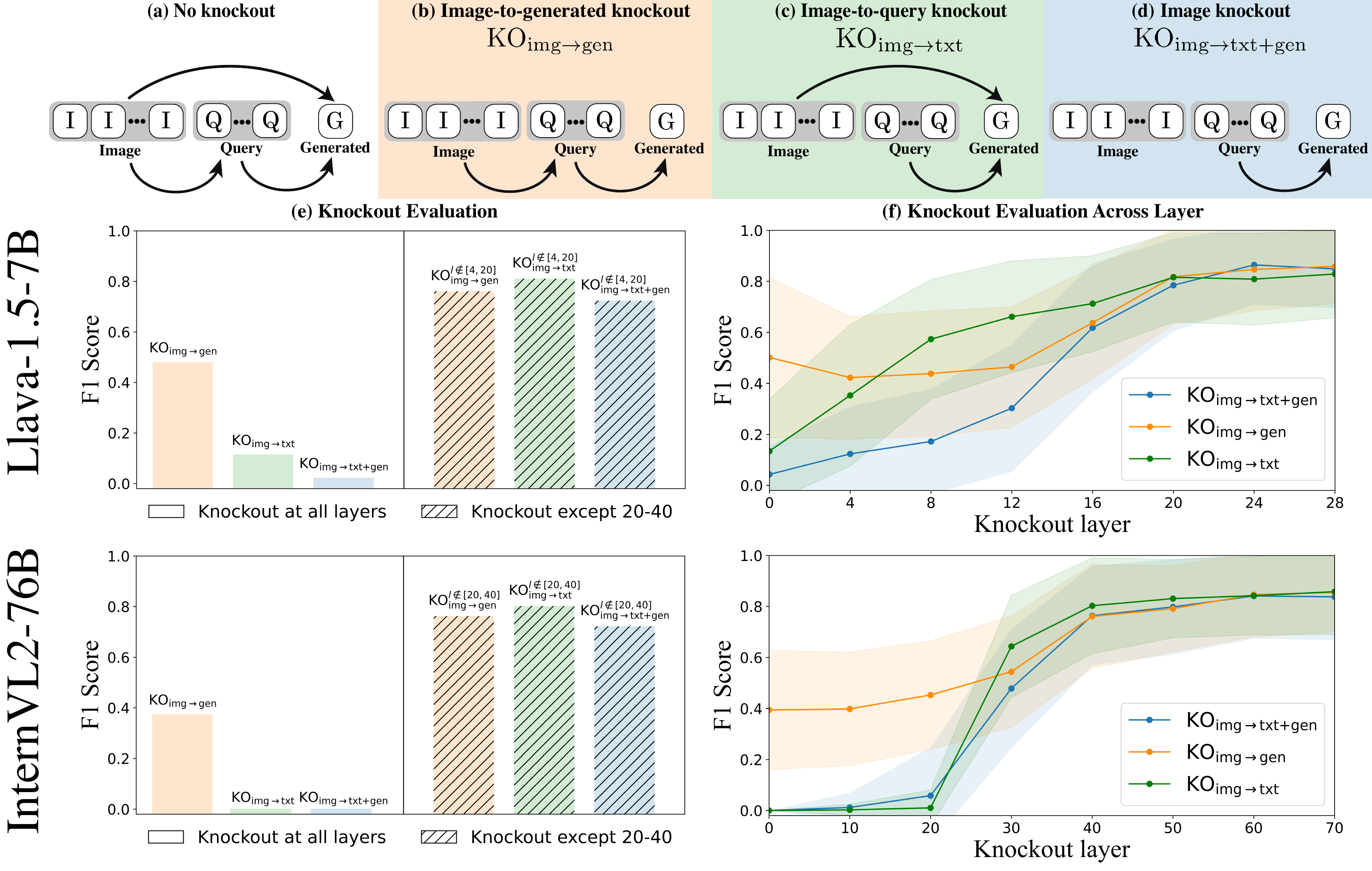} 
     \caption{ \textbf{Attention knockout on \llava~\cite{llava1_5} and \internvl~\cite{internvl1_5}:} (a) The VLM employs causal masking (Eq.~\ref{eq:causal_mask}), allowing generated and query tokens to gather information from image tokens, but not vice versa.  We analyze three knockout configurations: (b)~Image-to-generated $\KO{img}{gen}$: visual information flows to generated tokens only through query tokens, (c)~Image-to-query $\KO{img}{txt}$: blocks query tokens from accessing image information, and (d)~Image-to-others $\KO{img}{txt+gen}$: blocks image tokens from affecting all other tokens.
     (e) Evaluation of model responses (see Sec.~\ref{sec:llmjudge}) under each knockout configuration reveals that $\KO{img}{gen}$ achieves a 0.4 F1 score despite indirect image access, while $\KO{img}{txt}$ fails completely, demonstrating query tokens' essential role as global image descriptors. 
     (f)   We expand previous experiments by knocking out attention, starting from layer $l$. Results highlight a consistent rise in F1 scores in the mid-layers, suggesting their critical role in visual information processing.}
     \label{fig:knockout-sm}
     \afterfigure
\end{figure*}

\section{Annotating Fine-Grained Details}
\label{sec:fine_details_dataset_sm}
Sec.~\ref{sec:finegrained} explored how the model retrieves fine-grained visual information from image tokens.
For the purpose of this experiment, we defined a ``fine detail" as a concrete object that was spotted by the baseline VLM but was omitted under $KO_{\text{img}\rightarrow\text{gen}}$ knockout setting.
Sec.~\ref{sec:textual_tokens} showed that information conveying these objects is not being accumulated in the text query tokens, and the experiment in Sec.~\ref{sec:finegrained} was set to discover whether it comes by attending directly to image tokens.
To answer this question, we annotated fine details in images from the same subset of COCO images (Sec.~\ref{sec:coco-subset}).
We considered all false-negative details extracted during the LLM-as-a-judge evaluation for our visual-to-output knockout experiment of Fig.~\ref{fig:knockout-sm}(b) as candidate fine-grained visual details since the model was unable to describe them using the query text tokens alone.
Note that these details are not restricted to any pre-defined set of categories but rather follow an ``open vocabulary" setting where the details are defined based on analyzing differences in free-text image descriptions.
Furthermore, since the details are derived from a specific knockout experiment, different VLM models induce different lists of candidate details.
We further asked an LLM to associate each extracted detail with specific generated text tokens of the full description.
Given the textual description of the details in the images, we used text-guided segment anything model~\cite{zhang2024evf_sam} to create a binary mask localizing each detail in the image.
Finally, we manually inspected the extracted details and their masks and discarded details for which the masks did not match the textual description, were poorly localized or were of low quality, see examples in Fig.~\ref{fig:sam-fails}. 

After this manual selection, we were left with 231 annotated details in 68 images for \internvl, and 115 details in 57 images for \llava.
Each annotated detail comprises a segmentation mask, localizing it in \emph{image} space, and a short textual description, localizing it in the generated \emph{text}.

\section{\llava analysis}
\label{sec:llava_results}
\begin{figure}
     \centering
     \includegraphics[width=\linewidth]{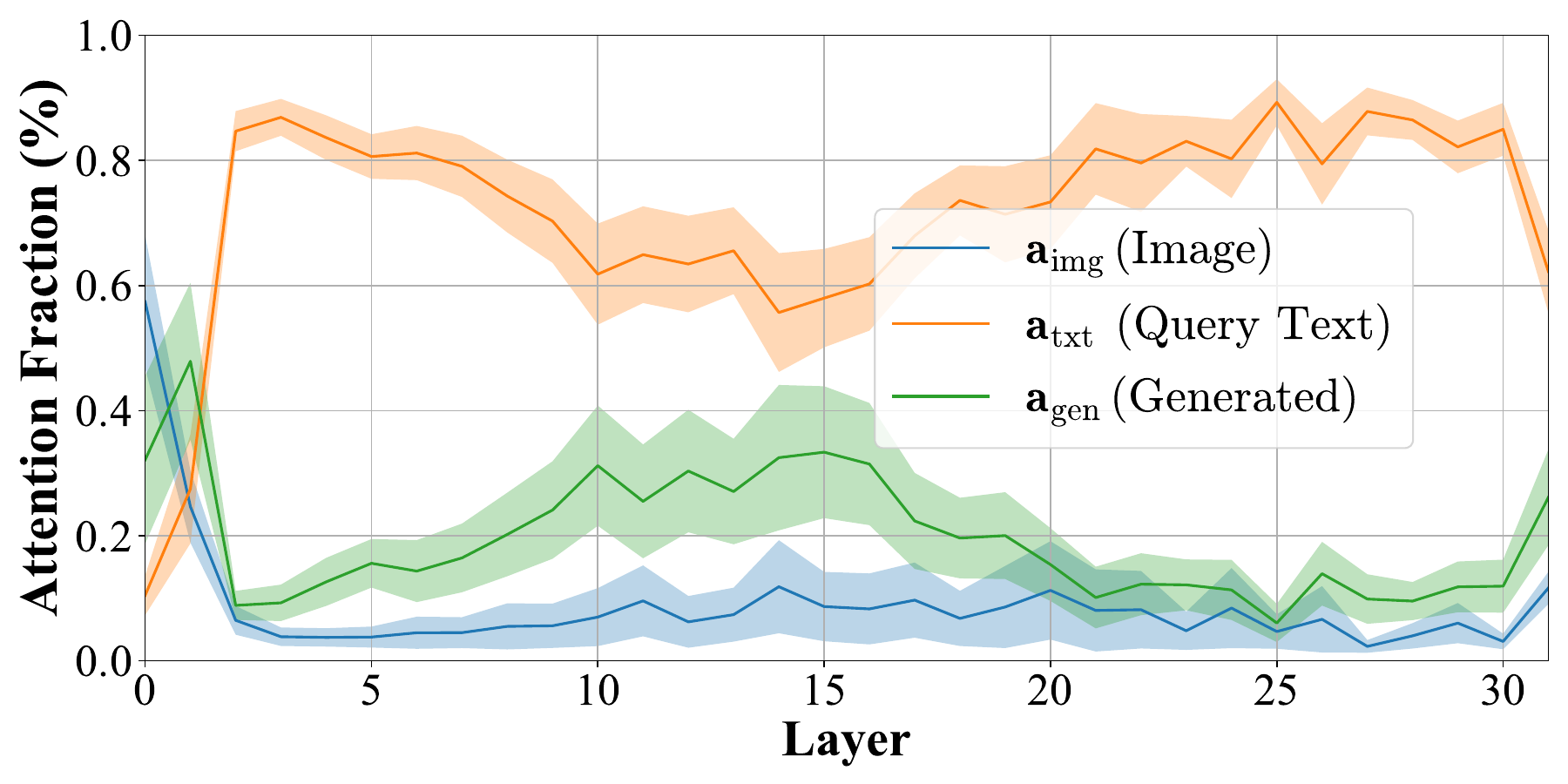} 
     \caption{ \textbf{\llava Fraction of attention to different token types:} We measure the relative amount by which the generated tokens attend to: image tokens (blue), query text tokens (orange), and the previously generated tokens in the sequence (green). We report the distribution of relative attention for a set of 80 images, averaged across attention heads and generated tokens for \llava.
     }
     \afterfigure
\label{fig:llava_attention_across_layers}
 \end{figure}

 \begin{figure}
    \centering
    \includegraphics[width=\linewidth]{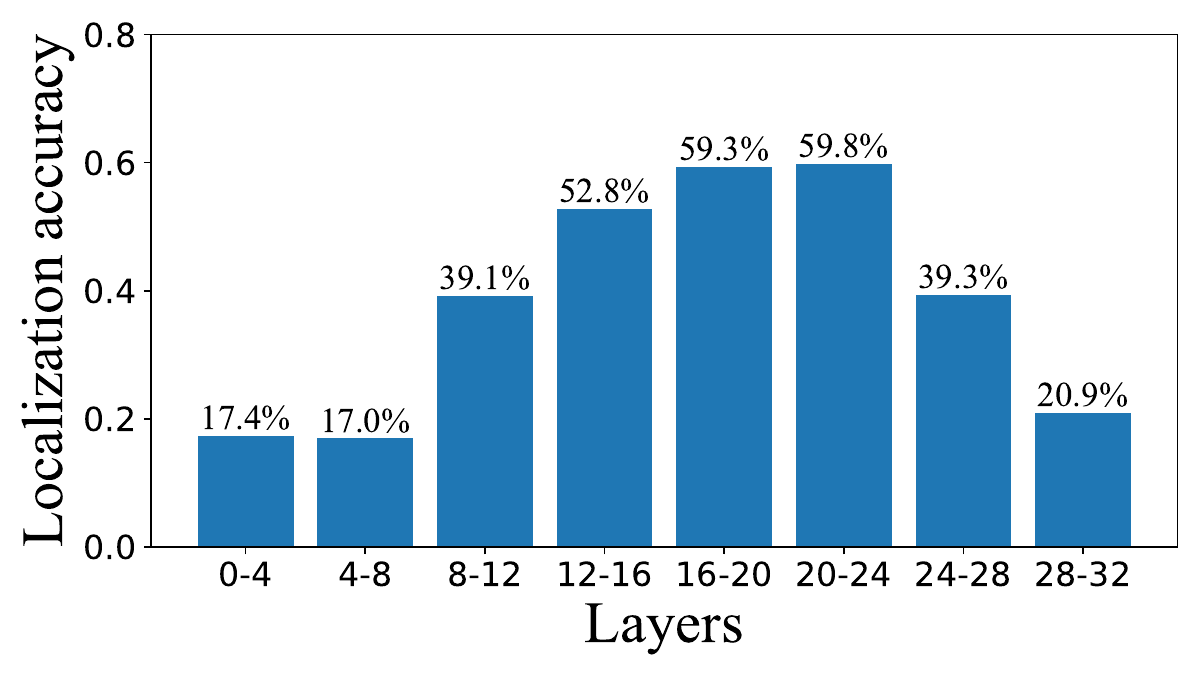}
    (a) \llava
    \includegraphics[width=\linewidth]{quantitative-accuracy-internvl2-76b.pdf}
    (b) \internvl
    \caption{\textbf{Object localization accuracy.}  
    We check if the attention of generated tokens associated with a specific object peak within one token distance from the pseudo ground truth object mask. We report the average accuracy across every four consecutive layers. 
    (a)~Results for \llava. (b)~Results for \internvl (presented in Fig.~\ref{fig:attn_to_details_quant} of the main paper and brought here for reference). The trend is similar for both models -- localization is done only in several middle layers.}
    \label{fig:llava-localization-quant}
    \vspace*{-4mm}
\end{figure}

\begin{figure}
    \centering
     \includegraphics[width=1\linewidth]{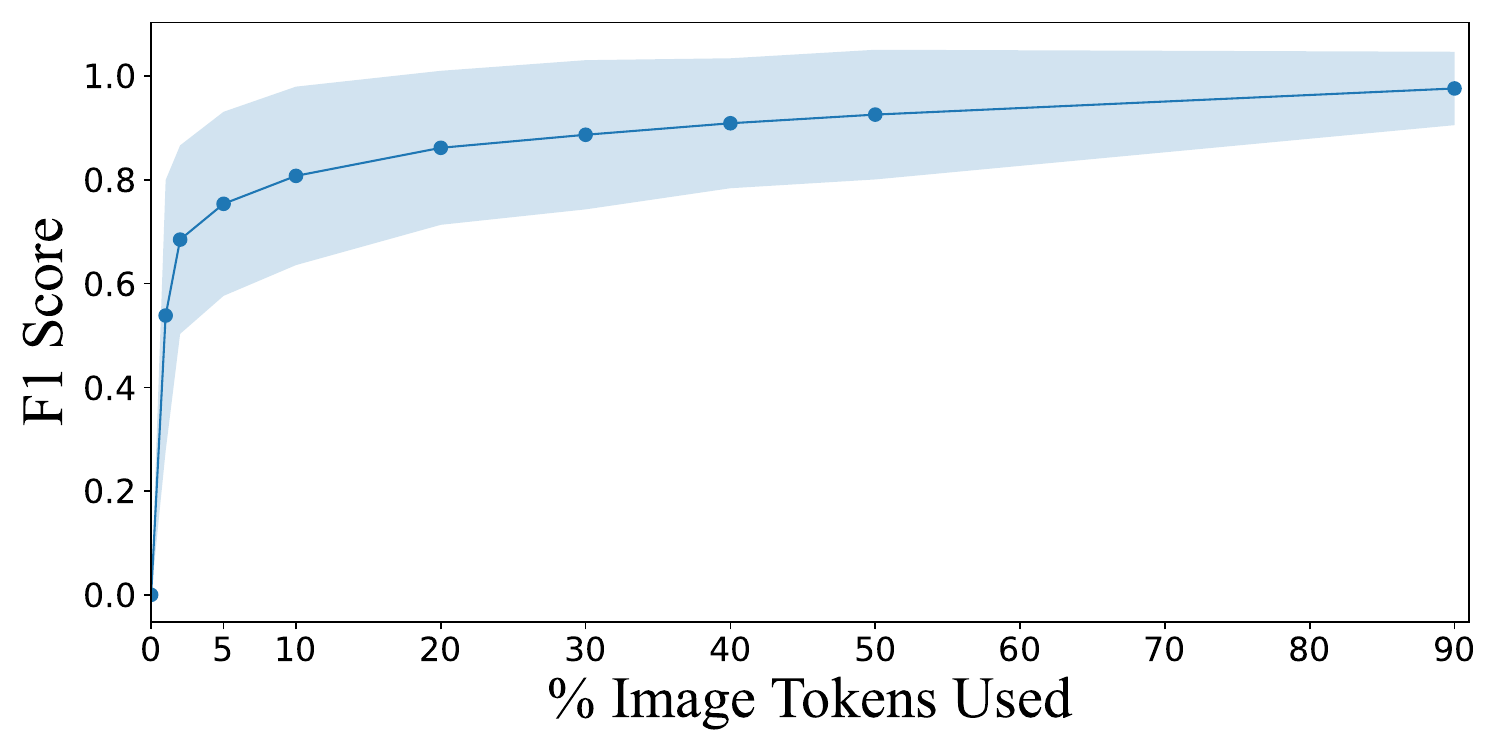} 
     \vspace{-0.8cm}     
     \caption{ \textbf{\llava Image tokens redundancy}: Similar to Fig.~\ref{fig:topk}, we evaluate the model performance, while letting the generated tokens access to only the top-k image tokens with the highest attention values. It does show a similar trend, where a small percentage of the tokens are enough to provide a high F1 score above 0.8. Moreover, we note that as \llava accepts only single-resolution patch, it has far less tokens, and 5\% in this case maps to only 13 tokens. }
     \label{fig:llava_f1_across_k}
     \afterfigure
\end{figure}

\begin{figure*}
    \hspace*{-0.025\linewidth}    
    \includegraphics[width=1.025\linewidth]{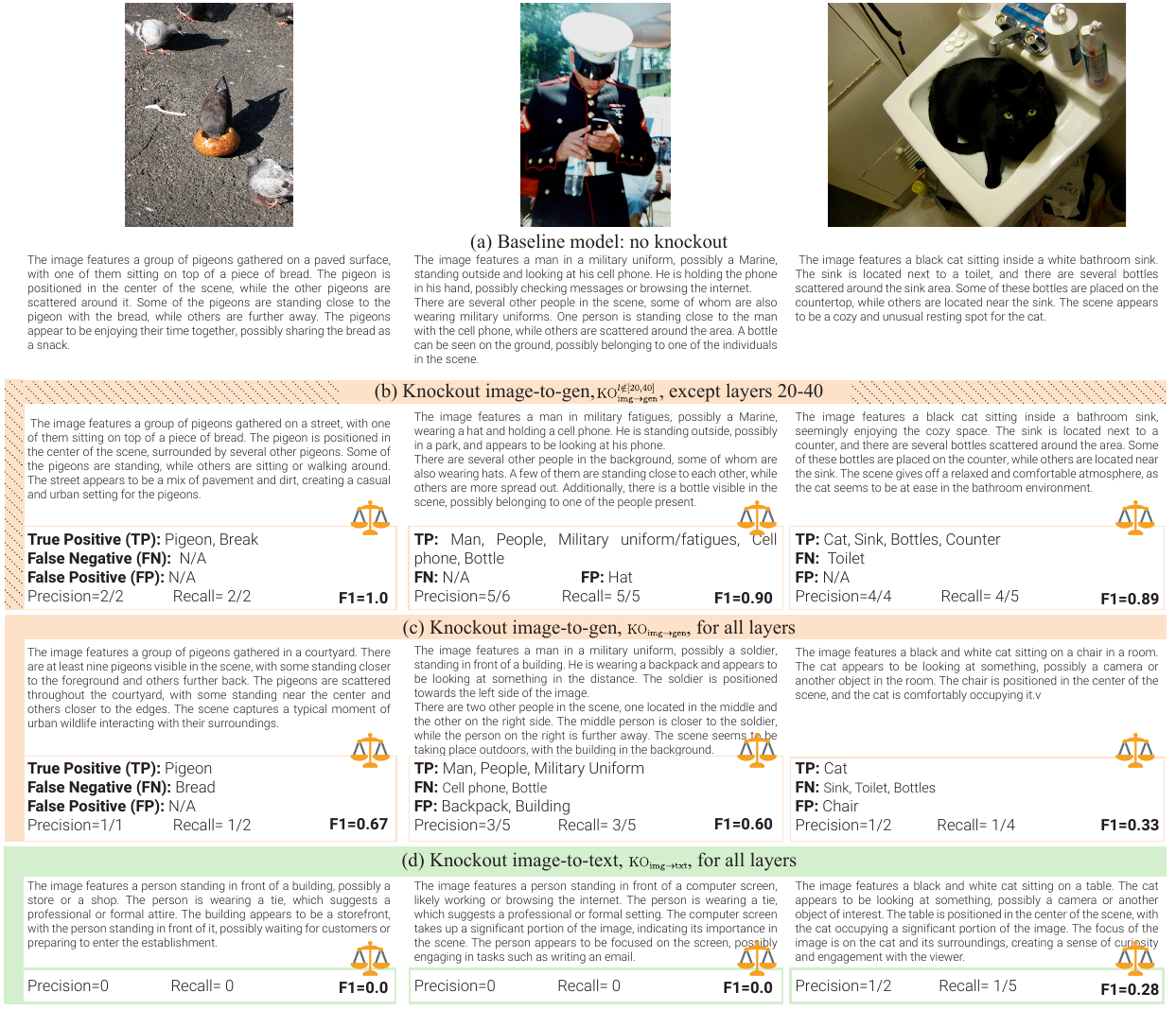}
    \vspace*{-5mm}
    \caption{\textbf{Qualitative results for knockout experiments on \llava:}  We use our LLM-as-a-judge protocol, \twemoji[scale=.375]{balance scale}, to compare the baseline VLM description of images~(a) to descriptions generated under various attention knockouts. (b)~Allowing generated tokens to attend to image tokens only in mid-layers 20-40, $\text{KO}_{\text{img}\rightarrow\text{gen}}^{l\notin\left[20, 40\right]}$, does not degrade the description significantly -- F1 scores are close to 1.0.
    (c)~Blocking attention between generated and image tokens for all layers, $\text{KO}_{\text{img}\rightarrow\text{gen}}$, results in loss of fine details, e.g., the bagel, smartphone or the toothpaste, and hallucinations, e.g., a black cap for the officer. Consequently, F1 scores are significantly lower -- around 0.45.
    (d)~When blocking attention between query text and image tokens for all layers, $\text{KO}_{\text{img}\rightarrow\text{txt}}$, the VLM is no longer able to describe the image
    -- F1=0.
    We note that LLM evaluation can be noisy, leading to slight inconsistencies in the identified objects across different comparisons. 
    For instance, in the rightmost examples, (b) and (c) show variations in the number of identified objects in the baseline (4 and 5).
    \label{fig:knockout-qualitative-llava}}
    \afterfigure
\end{figure*}

In this section we provide our analysis results on \llava-7B \cite{llava1_5}.

\paragraph{Attention Knockout Analysis}
We visualize in ~\ref{fig:llava_attention_across_layers} the fraction of attention towards each token  type: $\Tv$, $\Tt$, $\Tg$. It depicts a non-uniform flow of information across layers, as shown for \internvl in Fig.~\ref{fig:attention_across_layers}. 

We repeat our knockout experiments from Sec.~\ref{sec:analysis} on \llava, and provide the results in ~\ref{fig:knockout-sm} for both \llava and \internvl. We observe that all trends and observations from \internvl also occur in \llava: (a) the query tokens have an essential role as global image descriptors, (b) there is a special role for the mid-layers. Specifically, the mid-layers 4-20, which are only about 50\% of the layers, are responsible for most part of the information flow between the image and text modalities. We note both models exhibit such redundancy (25\% of the layers are sufficient in \internvl, 50\% for \llava), and we hypothesize the difference comes mainly from the fact that \llava is much smaller in parameter size.

\paragraph{Top-K Image Tokens Importance}
Finally, in corresponds to Fig.~\ref{fig:topk} in the main paper, we turn to validate if the visual tokens also exhibit a redundancy, when evaluating the model's performance while allowing only the top-k highest attended tokens to influence the generated tokens. Results are provided in Fig.~\ref{fig:llava_f1_across_k}, and indicates that for \llava a redundancy exists as well. However, it saturates slower, and we hypothesize it is due to the fact that \llava has much fewer visual tokens (256 vs 1600 on average for \internvl), a difference which stems from the multi-resolution encoding strategy of \internvl. Therefore, in \llava, using 5\% of the tokens is only 13 tokens, relative to 80 tokens in \internvl.

Qualitative results for the different knockout settings, on the same images used in the main paper at Fig.~\ref{fig:knockout-qualitative}, is provided in Fig.~\ref{fig:knockout-qualitative-llava}.

\paragraph{Fine-Grained Details Localized in Mid Layers}

\begin{figure*}
    \centering
    \vspace*{-5mm}
    \includegraphics[width=.975\linewidth]{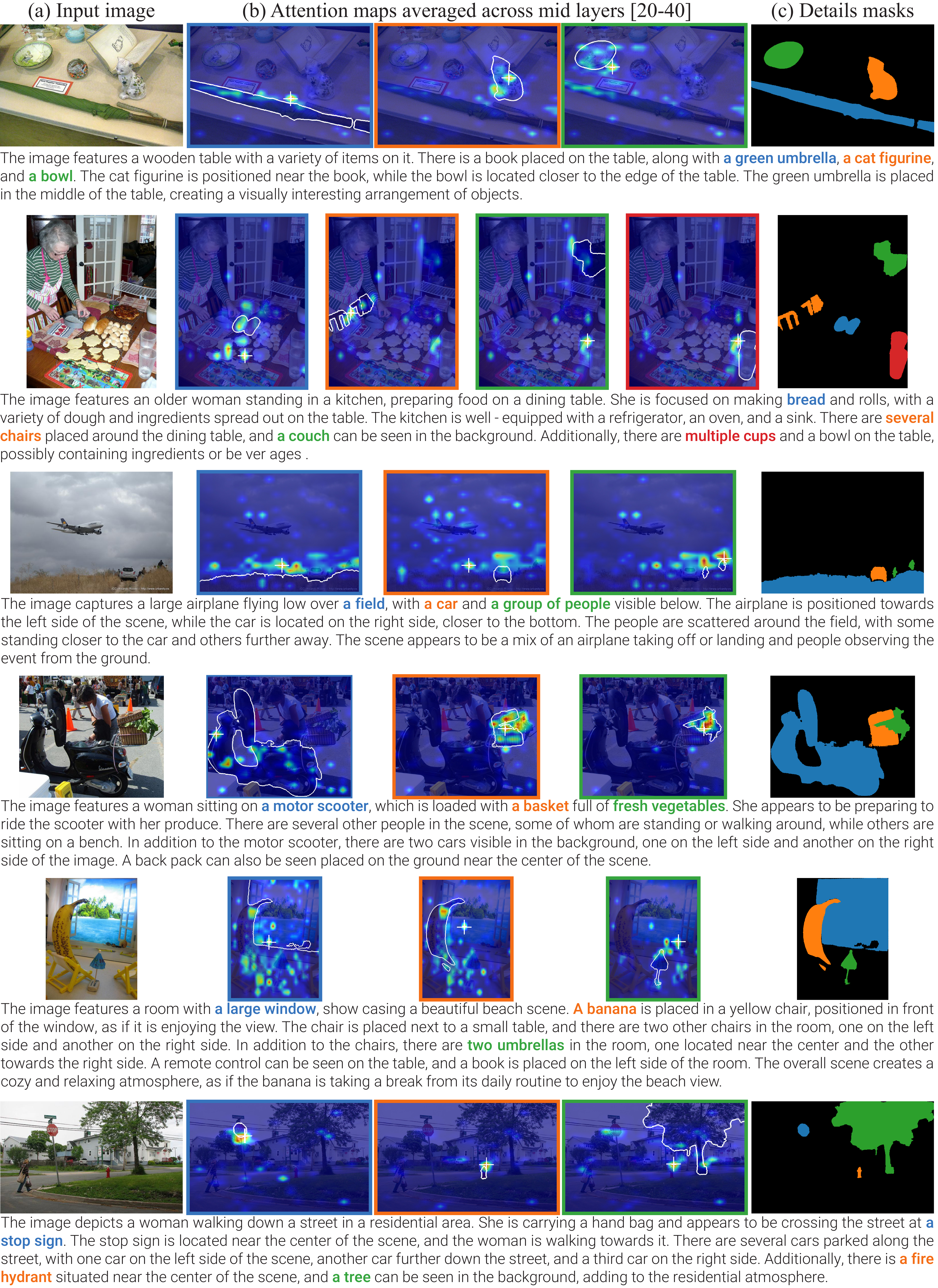}
    \caption{\textbf{Attending to objects:} Results for \llava. (a)~Input image. (b)~Average attention maps of the generated tokens associated with each object (marked in color in the generated text). (c)~Pseudo ground truth object masks, generated using SAM~\cite{kirillov2023sam, zhang2024evf_sam}. The peak of attention, marked by a white cross, aligns with the location of the object in the image, not as well as for \internvl.}
    \label{fig:attn-to-details-llava-sm}
\end{figure*}

Fig.~\ref{fig:attn-to-details-llava-sm} shows examples of the annotated details localized both in the image (segmentation mask) and in the generated text.
The aggregated attention maps of the mid-layers (layers 16-24 for \llava), corresponding to the generated text tokens, show good localization of the details in the image.

Additionally, we report localization accuracy for the annotated details in Fig.~\ref{fig:llava-localization-quant}.
The trend is similar for both models --
localization is done only in several middle layers.

\begin{figure*}
    \centering
    \includegraphics[width=\linewidth]{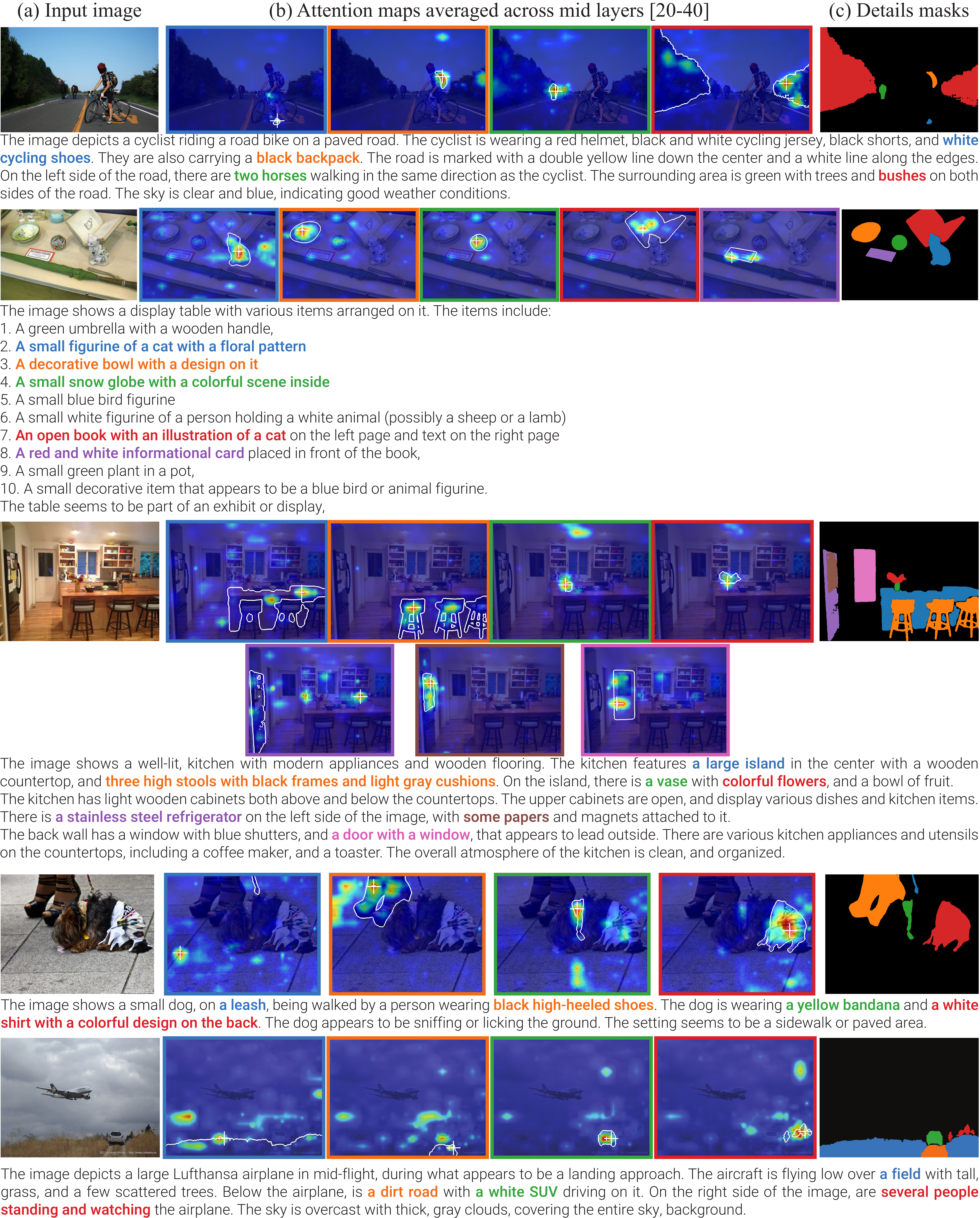}
    \caption{\textbf{Attending to objects:} Completing results for \internvl shown in Fig.~\ref{fig:attn_to_details} of the main paper. (a)~Input image. (b)~Average attention maps of the generated tokens associated with each object (marked in color in the generated text). (c)~Pseudo ground truth object masks, generated using SAM~\cite{kirillov2023sam, zhang2024evf_sam}. The peak of attention, marked by a white cross, aligns well with the location of the object in the image.}
    \label{fig:attn-to-details-internvl-sm}
\end{figure*}


\section{Images used for evaluation}
\label{sec:coco-subset}
All evaluations in the paper were conducted using a subset of 81 images from the COCO~\cite{lin2014coco} dataset.
Fig.~\ref{fig:coco-images} shows the photos selected and their IDs.
The images depict complex scenarios of various indoor and outdoor scenes with many fine details. To evaluate Image Reprompting (Sec.~\ref{sec:efficientvlm}), we used the MME dataset~\cite{fu2023mme}.



 %

%


\begin{figure*}
    \centering
     \includegraphics[width=1\textwidth]{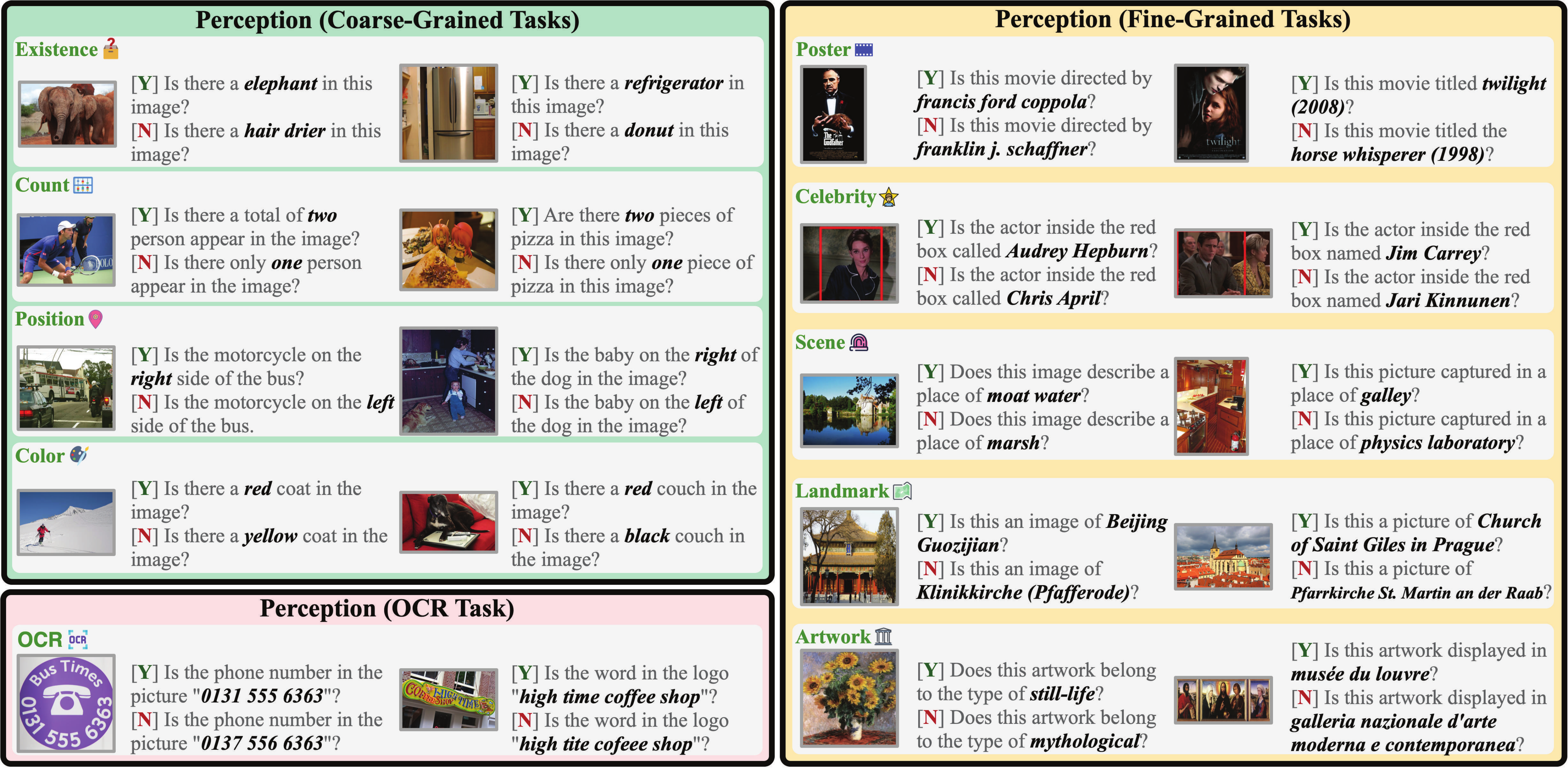}
     \captionof{figure}{\textbf{MME perception tasks:}  Illustration of the different tasks of the MME benchmark, taken from~\cite{fu2023mme} (cf Fig.~\ref{fig:attention_across_layers}). MME contains ten perception tasks. Each image is associated with two questions whose answers are marked yes [Y] or no [N], respectively. The instruction consists of a question followed by “Please answer yes or no”. Results over all subsets are provided in Tab.~\ref{table:mme_evaluation_combined}}
     \label{fig:mme_dataset}
\end{figure*}


\begin{figure*}[t]
    \centering
    \newlength{\ccimgh}
    \setlength{\ccimgh}{12.25mm}
    \begin{tabular}{c@{}c@{}c@{}c@{}c@{}c@{}c@{}c}
\includegraphics[height=\ccimgh]{000000480122.jpg} &
\includegraphics[height=\ccimgh]{000000256941.jpg} &
\includegraphics[height=\ccimgh]{000000414510.jpg} &
\includegraphics[height=\ccimgh]{000000245764.jpg} &
\includegraphics[height=\ccimgh]{000000307074.jpg} &
\includegraphics[height=\ccimgh]{000000261888.jpg} &
\includegraphics[height=\ccimgh]{000000246963.jpg} &
\includegraphics[height=\ccimgh]{000000123585.jpg} \\
{\footnotesize 480122} &
{\footnotesize 256941} &
{\footnotesize 414510} &
{\footnotesize 245764} &
{\footnotesize 307074} &
{\footnotesize 261888} &
{\footnotesize 246963} &
{\footnotesize 123585} \\
\includegraphics[height=\ccimgh]{000000211825.jpg} &
\includegraphics[height=\ccimgh]{000000122166.jpg} &
\includegraphics[height=\ccimgh]{000000555050.jpg} &
\includegraphics[height=\ccimgh]{000000232563.jpg} &
\includegraphics[height=\ccimgh]{000000114770.jpg} &
\includegraphics[height=\ccimgh]{000000037777.jpg} &
\includegraphics[height=\ccimgh]{000000441586.jpg} &
\includegraphics[height=\ccimgh]{000000293390.jpg} \\
{\footnotesize 211825} &
{\footnotesize 122166} &
{\footnotesize 555050} &
{\footnotesize 232563} &
{\footnotesize 114770} &
{\footnotesize 37777} &
{\footnotesize 441586} &
{\footnotesize 293390} \\
\includegraphics[height=\ccimgh]{000000045596.jpg} &
\includegraphics[height=\ccimgh]{000000520301.jpg} &
\includegraphics[height=\ccimgh]{000000291634.jpg} &
\includegraphics[height=\ccimgh]{000000292456.jpg} &
\includegraphics[height=\ccimgh]{000000123480.jpg} &
\includegraphics[height=\ccimgh]{000000017899.jpg} &
\includegraphics[height=\ccimgh]{000000550714.jpg} &
\includegraphics[height=\ccimgh]{000000266400.jpg} \\
{\footnotesize 45596} &
{\footnotesize 520301} &
{\footnotesize 291634} &
{\footnotesize 292456} &
{\footnotesize 123480} &
{\footnotesize 17899} &
{\footnotesize 550714} &
{\footnotesize 266400} \\
\includegraphics[height=\ccimgh]{000000078823.jpg} &
\includegraphics[height=\ccimgh]{000000069213.jpg} &
\includegraphics[height=\ccimgh]{000000146667.jpg} &
\includegraphics[height=\ccimgh]{000000228214.jpg} &
\includegraphics[height=\ccimgh]{000000134096.jpg} &
\includegraphics[height=\ccimgh]{000000480944.jpg} &
\includegraphics[height=\ccimgh]{000000251140.jpg} &
\includegraphics[height=\ccimgh]{000000554002.jpg} \\
{\footnotesize 78823} &
{\footnotesize 69213} &
{\footnotesize 146667} &
{\footnotesize 228214} &
{\footnotesize 134096} &
{\footnotesize 480944} &
{\footnotesize 251140} &
{\footnotesize 554002} \\
\includegraphics[height=\ccimgh]{000000023272.jpg} &
\includegraphics[height=\ccimgh]{000000343496.jpg} &
\includegraphics[height=\ccimgh]{000000066231.jpg} &
\includegraphics[height=\ccimgh]{000000125850.jpg} &
\includegraphics[height=\ccimgh]{000000241319.jpg} &
\includegraphics[height=\ccimgh]{000000323709.jpg} &
\includegraphics[height=\ccimgh]{000000492878.jpg} &
\includegraphics[height=\ccimgh]{000000529568.jpg} \\
{\footnotesize 23272} &
{\footnotesize 343496} &
{\footnotesize 66231} &
{\footnotesize 125850} &
{\footnotesize 241319} &
{\footnotesize 323709} &
{\footnotesize 492878} &
{\footnotesize 529568} \\
\includegraphics[height=\ccimgh]{000000191845.jpg} &
\includegraphics[height=\ccimgh]{000000508602.jpg} &
\includegraphics[height=\ccimgh]{000000274687.jpg} &
\includegraphics[height=\ccimgh]{000000440475.jpg} &
\includegraphics[height=\ccimgh]{000000543047.jpg} &
\includegraphics[height=\ccimgh]{000000501523.jpg} &
\includegraphics[height=\ccimgh]{000000052412.jpg} &
\includegraphics[height=\ccimgh]{000000096001.jpg} \\
{\footnotesize 191845} &
{\footnotesize 508602} &
{\footnotesize 274687} &
{\footnotesize 440475} &
{\footnotesize 543047} &
{\footnotesize 501523} &
{\footnotesize 52412} &
{\footnotesize 96001} \\
\includegraphics[height=\ccimgh]{000000267537.jpg} &
\includegraphics[height=\ccimgh]{000000426166.jpg} &
\includegraphics[height=\ccimgh]{000000298396.jpg} &
\includegraphics[height=\ccimgh]{000000284623.jpg} &
\includegraphics[height=\ccimgh]{000000058705.jpg} &
\includegraphics[height=\ccimgh]{000000134886.jpg} &
\includegraphics[height=\ccimgh]{000000579321.jpg} &
\includegraphics[height=\ccimgh]{000000101762.jpg} \\
{\footnotesize 267537} &
{\footnotesize 426166} &
{\footnotesize 298396} &
{\footnotesize 284623} &
{\footnotesize 58705} &
{\footnotesize 134886} &
{\footnotesize 579321} &
{\footnotesize 101762} \\
\includegraphics[height=\ccimgh]{000000226417.jpg} &
\includegraphics[height=\ccimgh]{000000357737.jpg} &
\includegraphics[height=\ccimgh]{000000240940.jpg} &
\includegraphics[height=\ccimgh]{000000472375.jpg} &
\includegraphics[height=\ccimgh]{000000530836.jpg} &
\includegraphics[height=\ccimgh]{000000569917.jpg} &
\includegraphics[height=\ccimgh]{000000575970.jpg} &
\includegraphics[height=\ccimgh]{000000560474.jpg} \\
{\footnotesize 226417} &
{\footnotesize 357737} &
{\footnotesize 240940} &
{\footnotesize 472375} &
{\footnotesize 530836} &
{\footnotesize 569917} &
{\footnotesize 575970} &
{\footnotesize 560474} \\
\includegraphics[height=\ccimgh]{000000157807.jpg} &
\includegraphics[height=\ccimgh]{000000109055.jpg} &
\includegraphics[height=\ccimgh]{000000185802.jpg} &
\includegraphics[height=\ccimgh]{000000070774.jpg} &
\includegraphics[height=\ccimgh]{000000222094.jpg} &
\includegraphics[height=\ccimgh]{000000289343.jpg} &
\includegraphics[height=\ccimgh]{000000245513.jpg} &
\includegraphics[height=\ccimgh]{000000061471.jpg} \\
{\footnotesize 157807} &
{\footnotesize 109055} &
{\footnotesize 185802} &
{\footnotesize 70774} &
{\footnotesize 222094} &
{\footnotesize 289343} &
{\footnotesize 245513} &
{\footnotesize 61471} \\ 
&
\includegraphics[height=\ccimgh]{000000494869.jpg} &
\includegraphics[height=\ccimgh]{000000493286.jpg} &
\includegraphics[height=\ccimgh]{000000259830.jpg} &
\includegraphics[height=\ccimgh]{000000309391.jpg} &
\includegraphics[height=\ccimgh]{000000005477.jpg} &
\includegraphics[height=\ccimgh]{000000425226.jpg} &
 \\
 &
{\footnotesize 494869} &
{\footnotesize 493286} &
{\footnotesize 259830} &
{\footnotesize 309391} &
{\footnotesize 5477} &
{\footnotesize 425226} &
 \\
 &
\includegraphics[height=\ccimgh]{000000093437.jpg} 
& \multicolumn{3}{c}{\includegraphics[height=\ccimgh]{000000527220.jpg}} &
\multicolumn{2}{c}{\includegraphics[height=\ccimgh]{000000096549.jpg}} \\
&
{\footnotesize 93437} 
&\multicolumn{3}{c}{{\footnotesize 527220}} &
\multicolumn{2}{c}{{\footnotesize 96549}} \\
    \end{tabular}
    \vspace*{-3mm}
    \caption{\textbf{Images selected for VLM inspection.} We used the following images from COCO~\cite{lin2014coco} depicting complex and varied scenes. 
    Beneath each image appears its COCO ID.}
    \afterfigure
    \label{fig:coco-images}
\end{figure*}

\end{document}

%% file: math_commands.tex
\usepackage{amsmath,amsfonts,bm}

\newcommand{\Tt}{\mathbf{T}_\text{txt}}
\newcommand{\Tv}{\mathbf{T}_\text{img}}
\newcommand{\Tg}{\mathbf{T}_\text{gen}}
\newcommand{\KO}[2]{\text{KO}_{\text{#1}\rightarrow\text{#2}}}

\def\eqref#1{equation~\ref{#1}}

\def\1{\bm{1}}

\DeclareMathAlphabet{\mathsfit}{\encodingdefault}{\sfdefault}{m}{sl}
\SetMathAlphabet{\mathsfit}{bold}{\encodingdefault}{\sfdefault}{bx}{n}



%% file: preamble.tex
\usepackage{multirow}
\usepackage{twemojis}

\usepackage[normalem]{ulem}
\usepackage{ifthen}
\usepackage{xspace}

\newcommand{\omri}[2][]{%
    \ifthenelse{ \equal{#1}{} }
        {\textcolor{blue}{(OK) #2}}
        {\textcolor{blue}{(OK) \sout{#1} #2\xspace{}}}
}

\newcommand{\shai}[2][]{%
    \ifthenelse{ \equal{#1}{} }
        {\textcolor{orange}{(SB) #2}}
        {\textcolor{orange}{(SB) \sout{#1} #2\xspace{}}}
}

\newcommand{\llava}{LLaVA-1.5\xspace}
\newcommand{\internvl}{InternVL2\xspace}

\newcommand{\afterfigure}{\vspace{-0.3cm}}

\newcommand{\myparagraph}[1]{\vspace{0.05cm}\noindent{\textbf{#1}}}